\documentclass{article} % For LaTeX2e
\usepackage{iclr2016_conference,times}
\usepackage{hyperref}
\usepackage{url}

\pdfoutput=1 %for posting to arXiv

\usepackage{graphicx}
\usepackage{subfigure}
\usepackage{caption}
\usepackage{booktabs}
\usepackage{multirow}

\usepackage{tabularx}
\usepackage{microtype}
\usepackage{dsfont}
\usepackage{bbm}
\usepackage{amsmath}

\title{Deep Compression: Compressing Deep Neural Networks with Pruning, Trained Quantization and Huffman Coding}

\author{Song Han \\
Stanford University, Stanford, CA 94305, USA \\
\texttt{songhan@stanford.edu} \\
\AND
Huizi Mao \\
Tsinghua University, Beijing, 100084, China \\
\texttt{mhz12@mails.tsinghua.edu.cn} \\
\AND
William J. Dally \\
Stanford University, Stanford, CA 94305, USA \\
NVIDIA, Santa Clara, CA 95050, USA\\
\texttt{dally@stanford.edu} \\}

\iclrfinalcopy % Uncomment for camera-ready version

\begin{document}

\maketitle

\begin{abstract}
Neural networks are both computationally intensive and memory intensive, making them difficult to deploy on embedded systems with limited hardware resources. To address this limitation, we introduce ``deep compression'', a three stage pipeline: pruning, trained quantization and Huffman coding, that work together to reduce the storage requirement of neural networks by $35\times$ to $49\times$ without affecting their accuracy. 
Our method first prunes the network by learning only the important connections. Next, we quantize the weights to enforce weight sharing, finally, we apply Huffman coding. After the first two steps we retrain the network to fine tune the remaining connections and the quantized centroids. Pruning, reduces the number of connections by $9\times$ to $13\times$;   Quantization then reduces the number of bits that represent each connection from 32 to 5.  On the ImageNet dataset, our method reduced the storage required by AlexNet by $\bf{35\times}$, from 240MB to 6.9MB, without loss of accuracy. Our method reduced the size of VGG-16 by $\bf{49\times}$ from 552MB to 11.3MB, again with no loss of accuracy. This allows fitting the model into on-chip SRAM cache rather than off-chip DRAM memory. Our compression method also facilitates the use of complex neural networks in mobile applications where application size and download bandwidth are constrained. Benchmarked on CPU, GPU and mobile GPU, compressed network has $3\times$ to $4\times$ layerwise speedup and $3\times$ to $7\times$ better energy efficiency.

\end{abstract}

\section{Introduction}

Deep neural networks have evolved to the state-of-the-art technique for computer vision tasks \citep{hinton12}\citep{simonyan2014very}. Though these neural networks are very powerful, the large number of weights consumes considerable storage and memory bandwidth. For example, the AlexNet Caffemodel is over 200MB, and the VGG-16 Caffemodel is over 500MB \citep{caffemodel}. This makes it difficult to deploy deep neural networks on mobile system.

First, for many mobile-first companies such as Baidu and Facebook, various apps are updated via different app stores, and they are very sensitive to the size of the binary files. For example, App Store has the restriction ``apps above 100 MB will not download until you connect to Wi-Fi''. As a result, a feature that increases the binary size by 100MB will receive much more scrutiny than one that increases it by 10MB. Although having deep neural networks running on mobile has many great features such as better privacy, less network bandwidth and real time processing, the large storage overhead prevents deep neural networks from being incorporated into mobile apps.

The second issue is energy consumption. Running large neural networks require a lot of memory bandwidth to fetch the weights and a lot of computation to do dot products--- which in turn consumes considerable energy.  
Mobile devices are battery constrained, making power hungry applications such as deep neural networks hard to deploy. 

Energy consumption is dominated by memory access. Under 45nm CMOS technology, a 32 bit floating point add consumes $\bf{0.9pJ}$, a 32bit SRAM cache access takes $\bf{5pJ}$, while a 32bit DRAM memory access takes $\bf{640pJ}$, which is 3 orders of magnitude of an add operation. Large networks do not fit in on-chip storage and hence require the more costly DRAM accesses. Running a 1 billion connection neural network, for example, at 20fps would require $(20Hz)(1G)(640pJ)=12.8W$ just for DRAM access - well beyond the power envelope of a typical mobile device.

\begin{figure}[t]
\centering
\vspace{-10pt}
\scalebox{1}[1]{\includegraphics[scale=0.55]{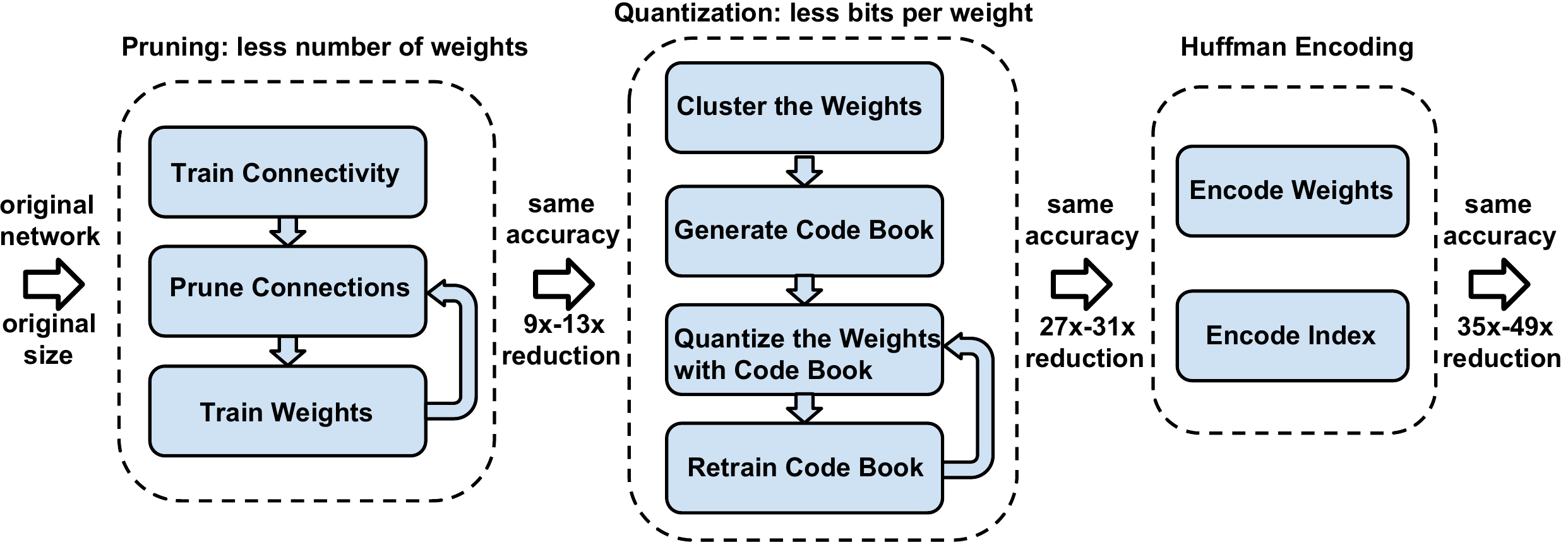}}
\caption{The three stage compression pipeline: pruning, quantization and Huffman coding. Pruning reduces the number of weights by $10\times$, while quantization further improves the compression rate: between $27\times$ and $31\times$. Huffman coding gives more compression: between $35\times$ and $49\times$. The compression rate already included the meta-data for sparse representation. The compression scheme doesn't incur any accuracy loss.}
\label{fig:pip}

\end{figure}

Our goal is to reduce the storage and energy required to run inference on such large networks so they can be deployed on mobile devices.
To achieve this goal, we present ``deep compression'': a three-stage pipeline (Figure \ref{fig:pip}) to reduce the storage required by neural network in a manner that preserves the original accuracy. First, we prune the networking by removing the redundant connections, keeping only the most informative connections. Next, the weights are quantized so that multiple connections share the same weight, thus only the codebook (effective weights) and the indices need to be stored. Finally, we apply Huffman coding to take advantage of the biased distribution of effective weights. 

Our main insight is that, pruning and trained quantization are able to compress the network without interfering each other, thus lead to surprisingly high compression rate. It makes the required storage so small (a few megabytes) that all weights can be cached on chip instead of going to off-chip DRAM which is energy consuming. Based on ``deep compression'', the EIE hardware accelerator~\cite{han2016eie} was later proposed that works on the compressed model, achieving significant speedup and energy efficiency improvement.

\section{Network Pruning}
Network pruning has been widely studied to compress CNN models. In early work, network pruning proved to be a valid way to reduce the network complexity and over-fitting \citep{lecun1989optimal, hanson1989comparing, hassibi1993second, strom1997phoneme}. Recently \citet{song_pruning} pruned state-of-the-art CNN models with no loss of accuracy. We build on top of that approach. As shown on the left side of Figure \ref{fig:pip}, we start by learning the connectivity via normal network training. Next, we prune the small-weight connections: all connections with weights below a threshold are removed from the network. Finally, we retrain the network to learn the final weights for the remaining sparse connections. Pruning reduced the number of parameters by $9\times$ and $13\times$ for AlexNet and VGG-16 model.

We store the sparse structure that results from pruning using compressed sparse row (CSR) or compressed sparse column (CSC) format, which requires $2a+n+1$ numbers, where $a$ is the number of non-zero elements and $n$ is the number of rows or columns.

\begin{figure}[t]
\centering
\vspace{-30pt}
\scalebox{1}[1]{\includegraphics[width=0.6\textwidth]{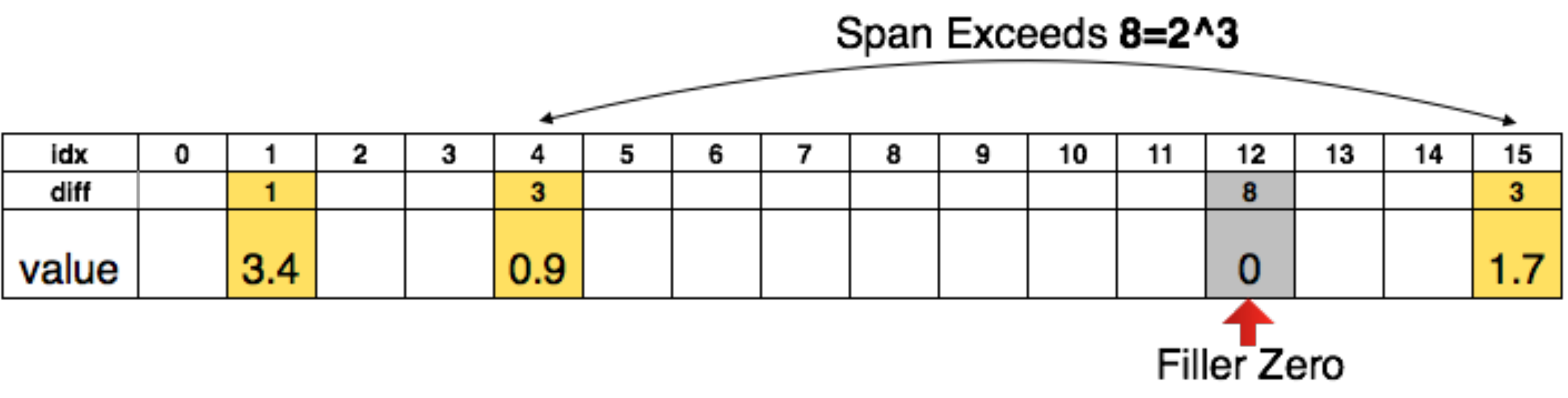}}
\caption{Representing the matrix sparsity with relative index. Padding filler zero to prevent overflow.}
\label{fig:jump}
\end{figure}

To compress further, we store the index difference instead of the absolute position, and encode this difference in 8 bits for conv layer and 5 bits for fc layer. When we need an index difference larger than the bound, we the zero padding solution shown in Figure \ref{fig:jump}: in case when the difference exceeds 8, the largest 3-bit (as an example) unsigned number, we add a filler zero. 

%%%%%%%%%%%%%%%%%%%%%%%%%%{Network Quantization}%%%%%%%%%%%%%%%%%%%%%%%%%%%%%%

\begin{figure}[t]
\centering
\scalebox{1}[1]{\includegraphics[scale=0.45]{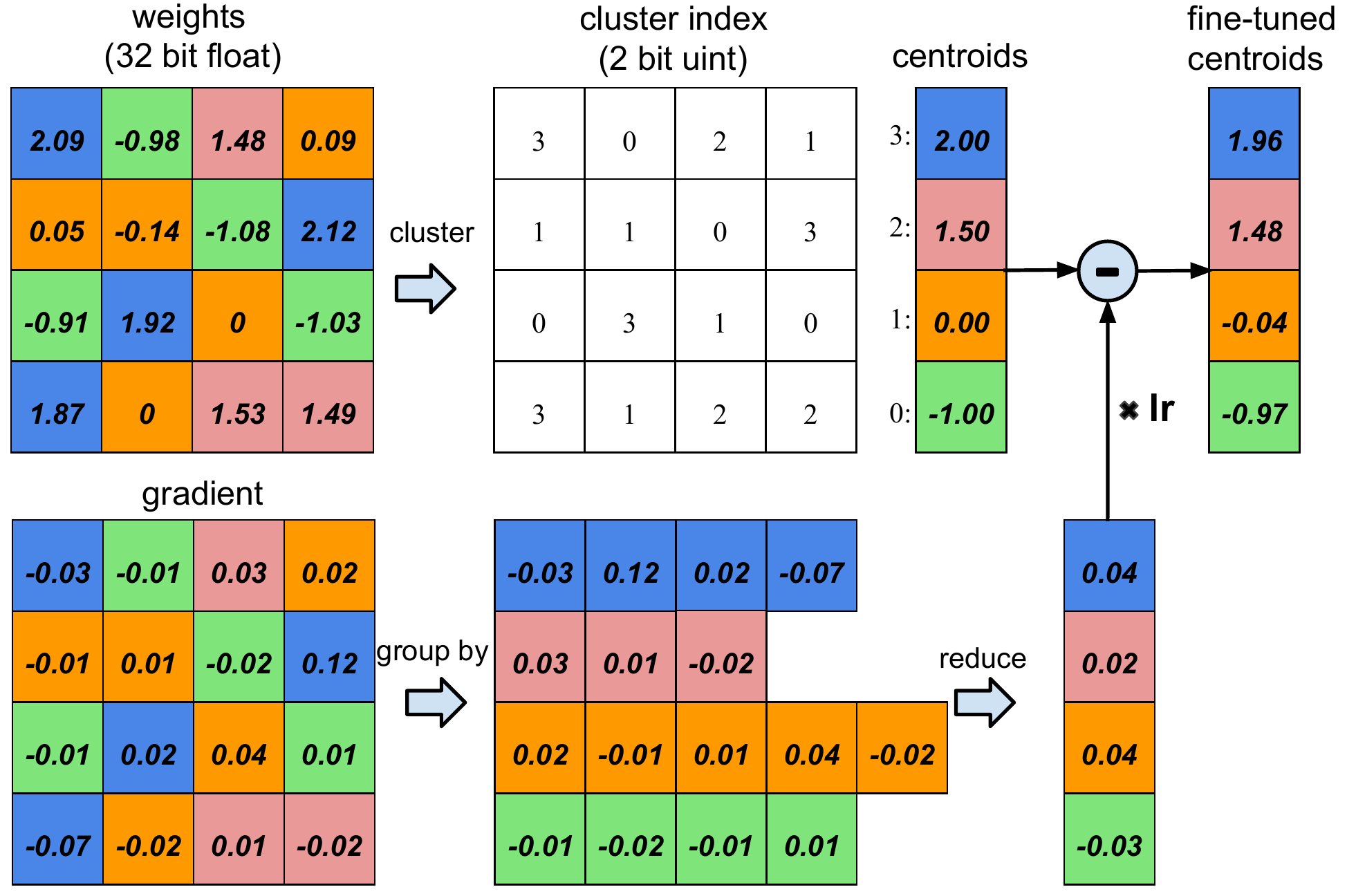}}
\caption{Weight sharing by scalar quantization (top) and centroids fine-tuning (bottom).}
\label{fig:theory}
\end{figure}

\section{Trained Quantization and Weight Sharing}
Network quantization and weight sharing further compresses the pruned network by reducing the number of bits required to represent each weight.  We limit the number of effective weights we need to store by having multiple connections share the same weight, and then fine-tune those shared weights.

Weight sharing is illustrated in Figure \ref{fig:theory}. Suppose we have a layer that has 4 input neurons and 4 output neurons, the weight is a $4\times 4$ matrix. On the top left is the $4\times 4$ weight matrix, and on the bottom left is the $4\times 4$ gradient matrix. The weights are quantized to 4 bins (denoted with 4 colors), all the weights in the same bin share the same value, thus for each weight, we then need to store only a small index into a table of shared weights. During update, all the gradients are grouped by the color and summed together, multiplied by the learning rate and subtracted from the shared centroids from last iteration. For pruned AlexNet, we are able to quantize to 8-bits (256 shared weights) for each CONV layers, and 5-bits (32 shared weights) for each FC layer without any loss of accuracy.

To calculate the compression rate, given $k$ clusters, we only need $log_{2}(k)$ bits to encode the index. In general, for a network with $n$ connections and each connection is represented with $b$ bits, constraining the connections to have only $k$ shared weights will result in a compression rate of:

\begin{equation}
r = \frac{nb}{nlog_{2}(k)+kb} 
\label{eq_save}
\end{equation}

For example, Figure \ref{fig:theory} shows the weights of a single layer neural network with four input units and four output units. There are $4\times 4 = 16$ weights originally but there are only $4$ shared weights: similar weights are grouped together to share the same value. Originally we need to store 16 weights each has 32 bits, now we need to store only 4 effective weights (blue, green, red and orange), each has 32 bits, together with 16 2-bit indices giving a compression rate of $16*32/(4*32+2*16) = 3.2$ 

%%%%%%%%%%%%%%%% Weight Sharing %%%%%%%%%%%%%%%%%%%%%%%%%%%%%%%%%%%%%%%%%%%%%%%
\subsection{Weight Sharing}
We use k-means clustering to identify the shared weights for each layer of a trained network, so that all the weights that fall into the same cluster will share the same weight. Weights are not shared across layers. We partition $n$ original weights $W=\{w_1, w_2, ..., w_n\}$ into $k$ clusters $C = \{c_1, c_2, ..., c_k\}$, $n\gg k$, so as to minimize the within-cluster sum of squares (WCSS):

\begin{equation}
\underset{C} {\operatorname{arg\min}}  \sum_{i=1}^{k} \sum_{ w \in c_i} \left|  w - c_i \right|^2 
\label{eq_save2}
\end{equation}

Different from HashNet \citep{chen2015compressing} where weight sharing is determined by a hash function before the networks sees any training data, our method determines weight sharing after a network is fully trained, so that the shared weights approximate the original network.

%%%%%%%%%%%%%%%% Initialization %%%%%%%%%%%%%%%%%%%%%%%%%%%%%%%%%%%%%%%%%%%%%%%

\begin{figure*}[t]
\centering
\vspace{-10pt}
\def \sz {1.84in}
\includegraphics[height=\sz]{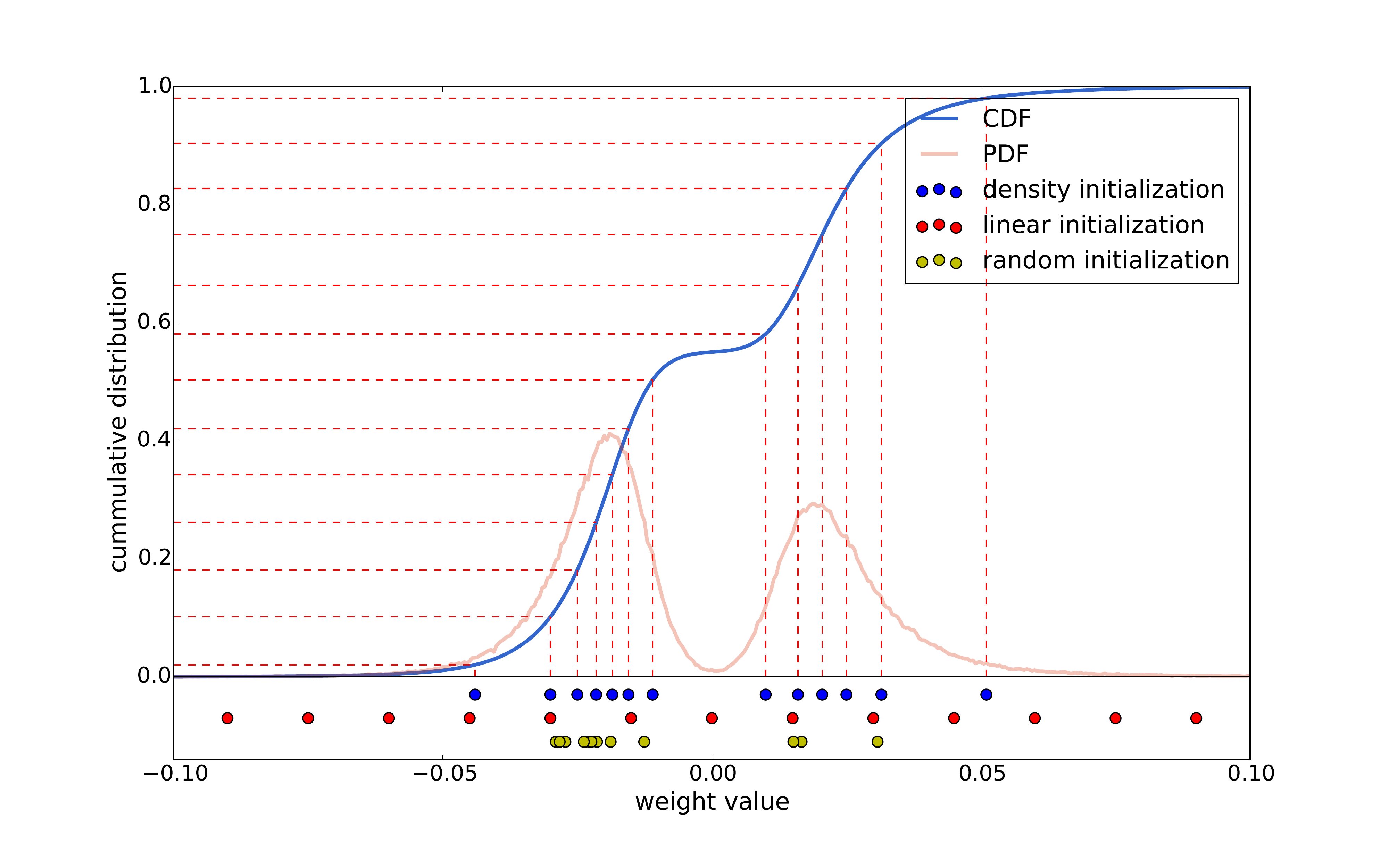}
\hspace{-2em}
\includegraphics[height=\sz]{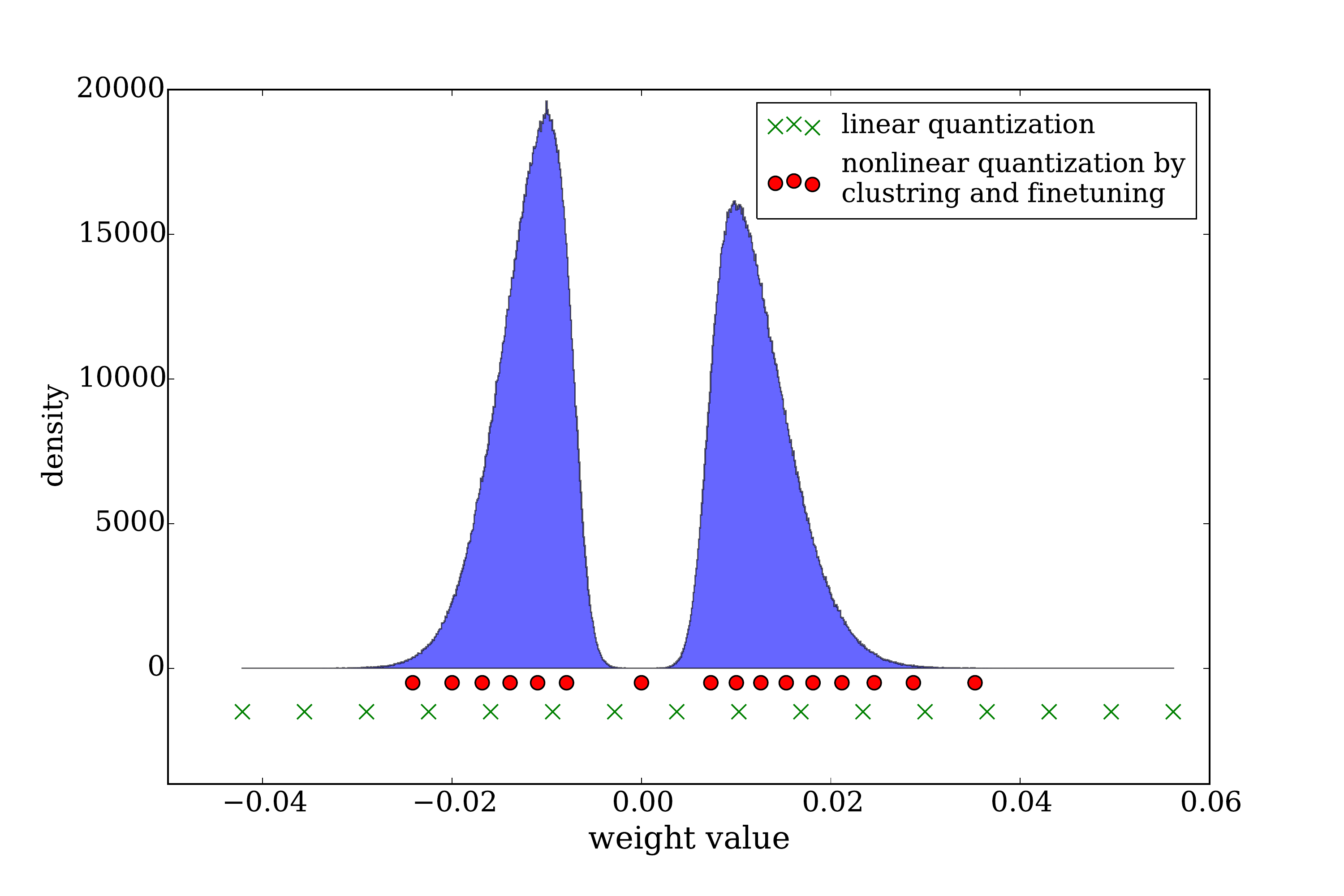}
\caption{Left: Three different methods for centroids initialization. Right: Distribution of weights (blue) and distribution of codebook before (green cross) and after fine-tuning (red dot).}
\label{fig:dist}

\end{figure*}

\subsection{Initialization of Shared Weights} 
Centroid initialization impacts the quality of clustering and thus affects the network's prediction accuracy. We examine three  initialization methods: Forgy(random), density-based, and linear initialization. In Figure \ref{fig:dist} we plotted the original weights' distribution of conv3 layer in AlexNet (CDF in blue, PDF in red). The weights forms a bimodal distribution after network pruning. On the bottom it plots the effective weights (centroids) with 3 different initialization methods (shown in blue, red and yellow). In this example, there are 13 clusters.

{\bf Forgy} (random) initialization randomly chooses k observations from the data set and uses these as the initial centroids. The initialized centroids are shown in yellow. Since there are two peaks in the bimodal distribution, Forgy method tend to concentrate around those two peaks. 

{\bf Density-based} initialization linearly spaces the CDF of the weights in the y-axis, then finds the horizontal intersection with the CDF, and finally finds the vertical intersection on the x-axis, which becomes a centroid, as shown in blue dots. This method makes the centroids denser around the two peaks, but more scatted than the Forgy method.

{\bf Linear} initialization linearly spaces the centroids between the [min, max] of the original weights. This initialization method is invariant to the distribution of the weights and is the most scattered compared with the former two methods.

Larger weights play a more important role than smaller weights \citep{song_pruning}, but there are fewer of these large weights. Thus for both Forgy initialization and density-based initialization, very few centroids have large absolute value which results in poor representation of these few large weights. Linear initialization does not suffer from this problem. The experiment section compares the accuracy of different initialization methods after clustering and fine-tuning, showing that linear initialization works best. 

%%%%%%%%%%%%%%%% Fine-tuning %%%%%%%%%%%%%%%%%%%%%%%%%%%%%%%%%%%%%%%%%%%%%%%%%%
\subsection{Feed-forward and Back-propagation}
The centroids of the one-dimensional k-means clustering are the shared weights. There is one level of indirection during feed forward phase and back-propagation phase looking up the weight table.   An index into the shared weight table is stored for each connection.  During back-propagation, the gradient for each shared weight is calculated and used to update
the shared weight. This procedure is shown in Figure \ref{fig:theory}.

We denote the loss by $\cal{L}$, the weight in the $i$th column and $j$th row by $W_{ij}$, the centroid index of element $W_{i,j}$ by $I_{ij}$, the $k$th centroid of the layer by $C_k$. By using the indicator function $\mathds{1}(.)$, the gradient of the centroids is calculated as:

\begin{equation}
\vspace{-15pt}
\frac{\partial \cal{L}}{\partial C_k} = \sum_{i,j}\frac{\partial  \cal{L}}{\partial  W_{ij}}\frac{\partial W_{ij}}{\partial C_k} = \sum_{i,j}\frac{\partial  \cal{L}}{\partial  W_{ij}}\mathds{1}({ I_{ij}=k })
\vspace{-3pt}
\end{equation}

%%%%%%%%%%%%%%%% Huffman %%%%%%%%%%%%%%%%%%%%%%%%%%%%%%%%%%%%%%%%%%%%%%%%%%%%%%
\vspace{3pt}
\section{Huffman coding}
\vspace{-2pt}
\begin{figure*}[t]
\vspace{-30pt}
\centering
\def \sz {1.45in}
\includegraphics[height=\sz]{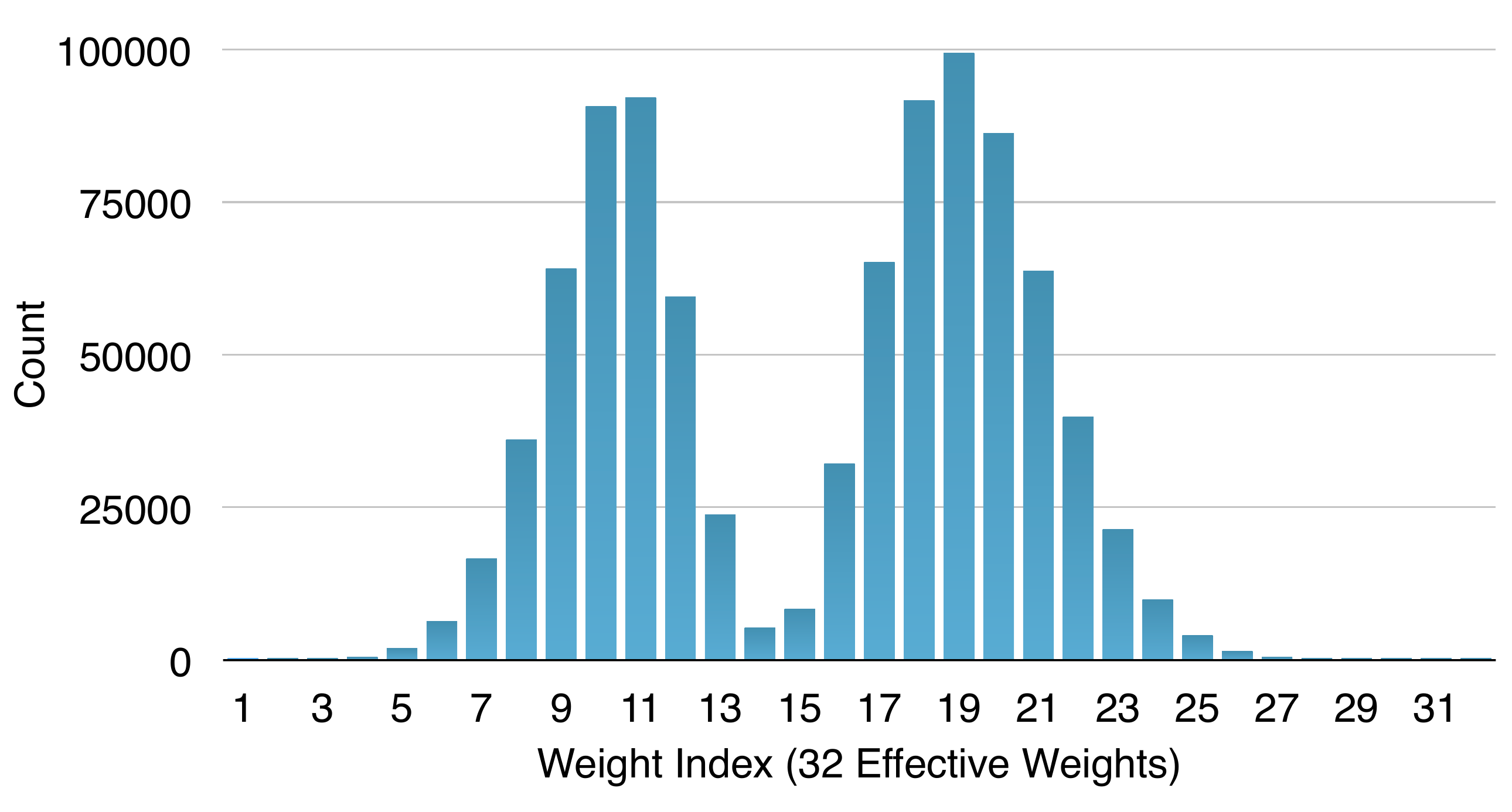}
\includegraphics[height=\sz]{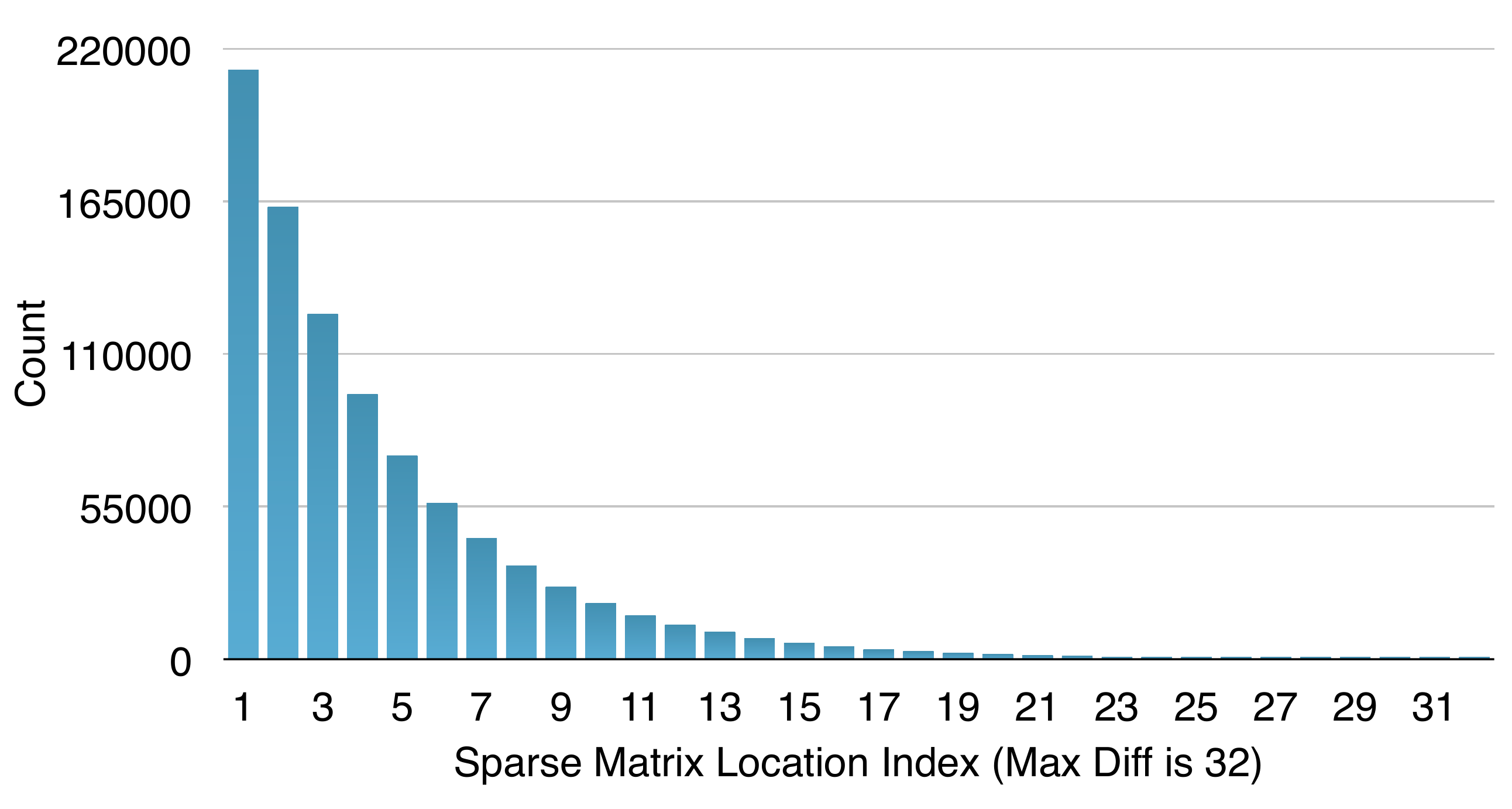}
\vspace{-8pt}
\caption{Distribution for weight (Left) and index (Right). The distribution is biased.}
\vspace{-10pt}
\label{fig:huffman}
\vspace{-5pt}
\end{figure*}

A Huffman code is an optimal prefix code commonly used for lossless data compression\citep{huffman}. It uses variable-length codewords to encode source symbols. The table is derived from the occurrence probability  for each symbol. More common symbols are represented with fewer bits. 

Figure \ref{fig:huffman} shows the probability distribution of quantized weights and the sparse matrix index of the last fully connected layer in AlexNet. Both distributions are biased: most of the quantized weights are distributed around the two peaks; the sparse matrix index difference are rarely above 20. Experiments show that Huffman coding these non-uniformly distributed values saves  $20\%-30\%$ of network storage.

%%%%%%%%%%%%%%%% Experiments %%%%%%%%%%%%%%%%%%%%%%%%%%%%%%%%%%%%%%%%%%%%%%%%%%
\vspace{-2pt}
\section{Experiments}
\vspace{-2pt}

We pruned, quantized, and Huffman encoded four networks: two on MNIST and two on ImageNet data-sets.
The network parameters and accuracy-\footnote{Reference model is from Caffe model zoo, accuracy is measured without data augmentation} before and after pruning are shown in Table \ref{table:results}. The compression pipeline saves network storage by $35\times$ to $49\times$ across different networks without loss of accuracy. The total size of AlexNet decreased from 240MB to 6.9MB, which is small enough to be put into on-chip SRAM, eliminating the need to store the model in energy-consuming DRAM memory. 

Training is performed with the Caffe framework \citep{jia2014caffe}. Pruning is implemented by adding a mask to the blobs to mask out the update of the pruned connections. Quantization and weight sharing are implemented by maintaining a codebook structure that stores the shared weight,  and group-by-index after calculating the gradient of each layer. Each shared weight is updated with all the gradients that fall into that bucket. Huffman coding doesn't require training and is implemented offline after all the fine-tuning is finished.

\begin{table}[t]
\centering
\vspace{-30pt}
\caption{The compression pipeline can save $35\times$ to $49\times$ parameter storage with no loss of accuracy.}
\begin{tabular}{lllll}
\hline
\multicolumn{1}{l|}{Network}                  & Top-1 Error & \multicolumn{1}{l|}{Top-5 Error} & \multicolumn{1}{l|}{Parameters} & \begin{tabular}[c]{@{}l@{}}Compress\\  Rate\end{tabular}  \\ \hline
\multicolumn{1}{l|}{LeNet-300-100 Ref}        & 1.64\%      & \multicolumn{1}{l|}{-}           & \multicolumn{1}{l|}{1070 KB} &        \\
\multicolumn{1}{l|}{LeNet-300-100 Compressed} & 1.58\%      & \multicolumn{1}{l|}{-}           & \multicolumn{1}{l|}{\bf{27 KB}} & $\bf{40\times}$ \\ \hline
\multicolumn{1}{l|}{LeNet-5 Ref}              & 0.80\%      & \multicolumn{1}{l|}{-}           & \multicolumn{1}{l|}{1720 KB} &        \\
\multicolumn{1}{l|}{LeNet-5 Compressed}       & 0.74\%      & \multicolumn{1}{l|}{-}           & \multicolumn{1}{l|}{\bf{44 KB}} & $\bf{39\times}$ \\ \hline
\multicolumn{1}{l|}{AlexNet Ref}              & 42.78\%     & \multicolumn{1}{l|}{19.73\%}     & \multicolumn{1}{l|}{240 MB} &        \\
\multicolumn{1}{l|}{AlexNet Compressed}       & 42.78\%     & \multicolumn{1}{l|}{19.70\%}     & \multicolumn{1}{l|}{\bf{6.9 MB}} & $\bf{35\times}$  \\ \hline
\multicolumn{1}{l|}{VGG-16 Ref}                & 31.50\%     & \multicolumn{1}{l|}{11.32\%}     & \multicolumn{1}{l|}{552 MB} &        \\
\multicolumn{1}{l|}{VGG-16 Compressed}         & 31.17\%     & \multicolumn{1}{l|}{10.91\%}     & \multicolumn{1}{l|}{\bf{11.3 MB}} & $\bf{49\times}$ \\ \hline
\end{tabular}
\label{table:results}
\end{table}

%%%%%%%%%%%%%%%%%%%%%%%%%%%   LeNet    %%%%%%%%%%%%%%%%%%%%%%%%%%%   
% lenet300-100
\begin{table}[t]
\centering
\caption{Compression statistics for LeNet-300-100. P: pruning, Q:quantization, H:Huffman coding.}
\resizebox{\textwidth}{!}{%
\begin{tabular}{l|llllllll}
\hline
Layer & \#Weights & \begin{tabular}[c]{@{}l@{}}Weights\%\\ (P)\end{tabular} & \begin{tabular}[c]{@{}l@{}}Weight\\ bits\\ (P+Q)\end{tabular} & \begin{tabular}[c]{@{}l@{}}Weight\\ bits\\ (P+Q+H)\end{tabular} & \begin{tabular}[c]{@{}l@{}}Index\\ bits\\ (P+Q)\end{tabular} & \begin{tabular}[c]{@{}l@{}}Index\\ bits\\ (P+Q+H)\end{tabular} & \begin{tabular}[c]{@{}l@{}}Compress\\ rate \\ (P+Q)\end{tabular} & \begin{tabular}[c]{@{}l@{}}Compress\\ rate\\ (P+Q+H)\end{tabular} \\ \hline
ip1 & 235K & 8\% & 6 & 4.4 & 5 & 3.7 & 3.1\% & 2.32\% \\
ip2 & 30K & 9\% & 6 & 4.4 & 5 & 4.3 & 3.8\% & 3.04\% \\
ip3 & 1K & 26\% & 6 & 4.3 & 5 & 3.2 & 15.7\% & 12.70\% \\ \hline
Total & 266K & 8\%($12\times$) & 6 & 5.1 & 5 & 3.7 & 3.1\% ($\bf{32\times}$) & 2.49\% ($\bf{40\times}$) \\ \hline
\end{tabular}
}
\label{Lenet300}
\end{table}

% lenet5
\begin{table}[t!]
\caption{Compression statistics for LeNet-5. P: pruning, Q:quantization, H:Huffman coding.}
\centering
\resizebox{\textwidth}{!}
{
\begin{tabular}{l|llllllll}
\hline
Layer & \#Weights & \begin{tabular}[c]{@{}l@{}}Weights\%\\ (P)\end{tabular} & \begin{tabular}[c]{@{}l@{}}Weight\\ bits\\ (P+Q)\end{tabular} & \begin{tabular}[c]{@{}l@{}}Weight\\ bits\\ (P+Q+H)\end{tabular} & \begin{tabular}[c]{@{}l@{}}Index\\ bits\\ (P+Q)\end{tabular} & \begin{tabular}[c]{@{}l@{}}Index\\ bits\\ (P+Q+H)\end{tabular} & \begin{tabular}[c]{@{}l@{}}Compress\\ rate \\ (P+Q)\end{tabular} & \begin{tabular}[c]{@{}l@{}}Compress\\ rate\\ (P+Q+H)\end{tabular} \\ \hline
conv1 & 0.5K & 66\% & 8 & 7.2 & 5 & 1.5 & 78.5\% & 67.45\% \\
conv2 & 25K & 12\% & 8 & 7.2 & 5 & 3.9 & 6.0\% & 5.28\% \\ \hline
ip1 & 400K & 8\% & 5 & 4.5 & 5 & 4.5 & 2.7\% & 2.45\% \\
ip2 & 5K & 19\% & 5 & 5.2 & 5 & 3.7 & 6.9\% & 6.13\% \\ \hline
Total & 431K & 8\%($12\times$) & 5.3 & 4.1 & 5 & 4.4 & 3.05\% ($\bf{33\times}$) & 2.55\% ($\bf{39\times}$)  \\ \hline
\end{tabular}
}
\label{Lenet5}
\end{table}

\subsection{LeNet-300-100 and LeNet-5 on MNIST}

We first experimented on MNIST dataset with LeNet-300-100 and LeNet-5 network \citep{lecun1998gradient}. LeNet-300-100 is a fully connected network with two hidden layers, with 300 and 100 neurons each, which achieves 1.6\% error rate on Mnist. LeNet-5 is a convolutional network that has two convolutional layers and two fully connected layers, which  achieves 0.8\% error rate on Mnist.  Table \ref{Lenet300} and table \ref{Lenet5} show the statistics of the compression pipeline. The compression rate includes the overhead of the codebook and sparse indexes. Most of the saving comes from pruning and quantization (compressed $32\times$), while Huffman coding gives a marginal gain (compressed $40\times$)

%%%%%%%%%%%%%%%%%%%%%%%%%%%   AlexNet    %%%%%%%%%%%%%%%%%%%%%%%%%%%   

\subsection{AlexNet on ImageNet}
We further examine the performance of Deep Compression on the ImageNet ILSVRC-2012 dataset, which has 1.2M training examples and 50k validation examples. We use the AlexNet Caffe model as the reference model, which has 61 million parameters and achieved a top-1 accuracy of 57.2\% and a top-5 accuracy of 80.3\%. Table \ref{Alexnet} shows that AlexNet can be compressed to $2.88\%$ of its original size without impacting accuracy. There are 256 shared weights in each CONV layer, which are encoded with 8 bits, and 32 shared weights in each FC layer, which are encoded with only 5 bits. The relative sparse index is encoded with 4 bits.  Huffman coding compressed additional 22\%, resulting in $35\times$ compression in total.

% alexnet table
\begin{table}[t]
\centering
\vspace{-30pt}
\caption{Compression statistics for AlexNet. P: pruning, Q: quantization, H:Huffman coding.}
\resizebox{\textwidth}{!}{%
\begin{tabular}{l|llllllll}
\hline
Layer & \#Weights & \begin{tabular}[c]{@{}l@{}}Weights\%\\ (P)\end{tabular} & \begin{tabular}[c]{@{}l@{}}Weight\\ bits\\ (P+Q)\end{tabular} & \begin{tabular}[c]{@{}l@{}}Weight\\ bits\\ (P+Q+H)\end{tabular} & \begin{tabular}[c]{@{}l@{}}Index\\ bits\\ (P+Q)\end{tabular} & \begin{tabular}[c]{@{}l@{}}Index\\ bits\\ (P+Q+H)\end{tabular} & \begin{tabular}[c]{@{}l@{}}Compress\\ rate \\ (P+Q)\end{tabular} & \begin{tabular}[c]{@{}l@{}}Compress\\ rate\\ (P+Q+H)\end{tabular} \\ \hline
conv1 & 35K & 84\% & 8 & 6.3 & 4 & 1.2 & 32.6\% & 20.53\% \\
conv2 & 307K & 38\% & 8 & 5.5 & 4 & 2.3 & 14.5\% & 9.43\% \\
conv3 & 885K & 35\% & 8 & 5.1 & 4 & 2.6 & 13.1\% & 8.44\% \\
conv4 & 663K & 37\% & 8 & 5.2 & 4 & 2.5 & 14.1\% & 9.11\% \\
conv5 & 442K & 37\% & 8 & 5.6 & 4 & 2.5 & 14.0\% & 9.43\% \\ \hline
fc6 & 38M & 9\% & 5 & 3.9 & 4 & 3.2 & 3.0\% & 2.39\% \\
fc7 & 17M & 9\% & 5 & 3.6 & 4 & 3.7 & 3.0\% & 2.46\% \\
fc8 & 4M & 25\% & 5 & 4 & 4 & 3.2 & 7.3\% & 5.85\% \\ \hline
Total & 61M & 11\%($9\times$) & 5.4 & 4 & 4 & 3.2 & 3.7\% ($\bf{27\times}$) & 2.88\% ($\bf{35\times}$) \\ \hline
\end{tabular}
}
\label{Alexnet}
\end{table}

%%%%%%%%%%%%%%%%%%%%%%%%%%%%%%%%%%%%  VGG  %%%%%%%%%%%%%%%%%%%%%%%%%%%%%%%%%%%%
\subsection{VGG-16 on ImageNet}
With promising results on AlexNet, we also looked at a larger, more recent network, VGG-16 \citep{simonyan2014very}, on the same ILSVRC-2012 dataset.  VGG-16 has far more convolutional layers but still only three fully-connected layers.  Following a similar methodology, we aggressively compressed both convolutional and fully-connected layers to realize a significant reduction in the number of effective weights, shown in Table\ref{VGGNet}.

\begin{table}[t]
\centering
\caption{Compression statistics for VGG-16. P: pruning, Q:quantization, H:Huffman coding.}
\resizebox{\textwidth}{!}{%
\begin{tabular}{l|llllllll}
\hline
Layer & \#Weights & \begin{tabular}[c]{@{}l@{}}Weights\%\\ (P)\end{tabular} & \begin{tabular}[c]{@{}l@{}}Weigh\\ bits\\ (P+Q)\end{tabular} & \begin{tabular}[c]{@{}l@{}}Weight\\ bits\\ (P+Q+H)\end{tabular} & \begin{tabular}[c]{@{}l@{}}Index\\ bits\\ (P+Q)\end{tabular} & \begin{tabular}[c]{@{}l@{}}Index\\ bits\\ (P+Q+H)\end{tabular} & \begin{tabular}[c]{@{}l@{}}Compress\\ rate \\ (P+Q)\end{tabular} & \begin{tabular}[c]{@{}l@{}}Compress\\ rate\\ (P+Q+H)\end{tabular} \\ \hline
conv1\_1 & 2K & 58\% & 8 & 6.8 & 5 & 1.7 & 40.0\% & 29.97\% \\
conv1\_2 & 37K & 22\% & 8 & 6.5 & 5 & 2.6 & 9.8\% & 6.99\% \\
conv2\_1 & 74K & 34\% & 8 & 5.6 & 5 & 2.4 & 14.3\% & 8.91\% \\
conv2\_2 & 148K & 36\% & 8 & 5.9 & 5 & 2.3 & 14.7\% & 9.31\% \\
conv3\_1 & 295K & 53\% & 8 & 4.8 & 5 & 1.8 & 21.7\% & 11.15\% \\
conv3\_2 & 590K & 24\% & 8 & 4.6 & 5 & 2.9 & 9.7\% & 5.67\% \\
conv3\_3 & 590K & 42\% & 8 & 4.6 & 5 & 2.2 & 17.0\% & 8.96\% \\
conv4\_1 & 1M & 32\% & 8 & 4.6 & 5 & 2.6 & 13.1\% & 7.29\% \\
conv4\_2 & 2M & 27\% & 8 & 4.2 & 5 & 2.9 & 10.9\% & 5.93\% \\
conv4\_3 & 2M & 34\% & 8 & 4.4 & 5 & 2.5 & 14.0\% & 7.47\% \\
conv5\_1 & 2M & 35\% & 8 & 4.7 & 5 & 2.5 & 14.3\% & 8.00\% \\
conv5\_2 & 2M & 29\% & 8 & 4.6 & 5 & 2.7 & 11.7\% & 6.52\% \\
conv5\_3 & 2M & 36\% & 8 & 4.6 & 5 & 2.3 & 14.8\% & 7.79\% \\ \hline
fc6 & 103M & 4\% & 5 & 3.6 & 5 & 3.5 & 1.6\% & 1.10\% \\
fc7 & 17M & 4\% & 5 & 4 & 5 & 4.3 & 1.5\% & 1.25\% \\
fc8 & 4M & 23\% & 5 & 4 & 5 & 3.4 & 7.1\% & 5.24\% \\ \hline
Total & 138M & 7.5\%($13\times$) & 6.4 & 4.1 & 5 & 3.1 & 3.2\% ($\bf{31\times}$) & 2.05\% ($\bf{49\times}$) \\ \hline
\end{tabular}
}
\vspace{-10pt}
\label{VGGNet}
\end{table}

The VGG16 network as a whole has been compressed by $49\times$.   Weights in the CONV layers are represented with 8 bits, and FC layers use 5 bits, which does not impact the accuracy.
The two largest fully-connected layers can each be pruned to less than 1.6\% of their original size.  
This reduction is critical for real time image processing, where there is little reuse of these layers across images (unlike batch processing). This is also critical for fast object detection algorithms where one CONV pass is used by many FC passes. 
The reduced layers will fit in an on-chip SRAM and have modest bandwidth requirements.
Without the reduction, the bandwidth requirements are prohibitive.

\vspace{-5pt}
\section{Discussions}
\vspace{-5pt}
%%%%%%%%%%%%%%%%%%%%%%  1. Working Together  %%%%%%%%%%%%%%%%%%%%%%
\subsection{Pruning and Quantization Working Together} 

\begin{figure}[t]
\centering
\vspace{-30pt}
\scalebox{1}[1]{\includegraphics[scale=0.45]{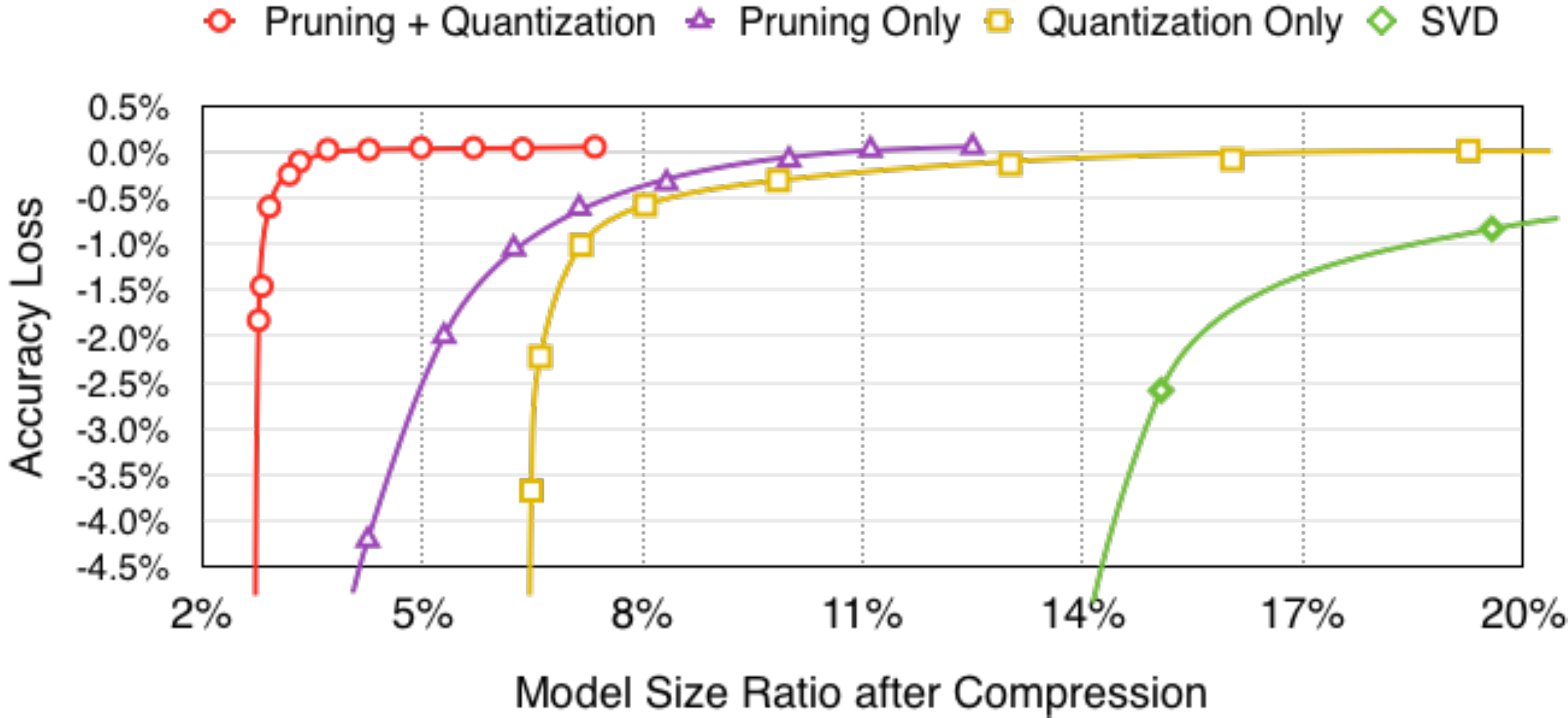}}
\caption{Accuracy v.s. compression rate under different compression methods. Pruning and quantization works best when combined.}
\vspace{-5pt}
\label{fig:acc}
\end{figure}

Figure \ref{fig:acc} shows the accuracy at different compression rates for pruning and quantization together or individually. When working individually, as shown in the purple and yellow lines, accuracy of pruned network begins to drop significantly when compressed below 8\% of its original size; accuracy of quantized network also begins to drop significantly when compressed below 8\% of its original size.  But when combined, as shown in the red line, the network can be compressed to 3\% of original size with no loss of accuracy.
On the far right side compared the result of SVD, which is inexpensive but has a poor compression rate.

\begin{figure*}[t]
\centering
\includegraphics[width=0.329\textwidth]{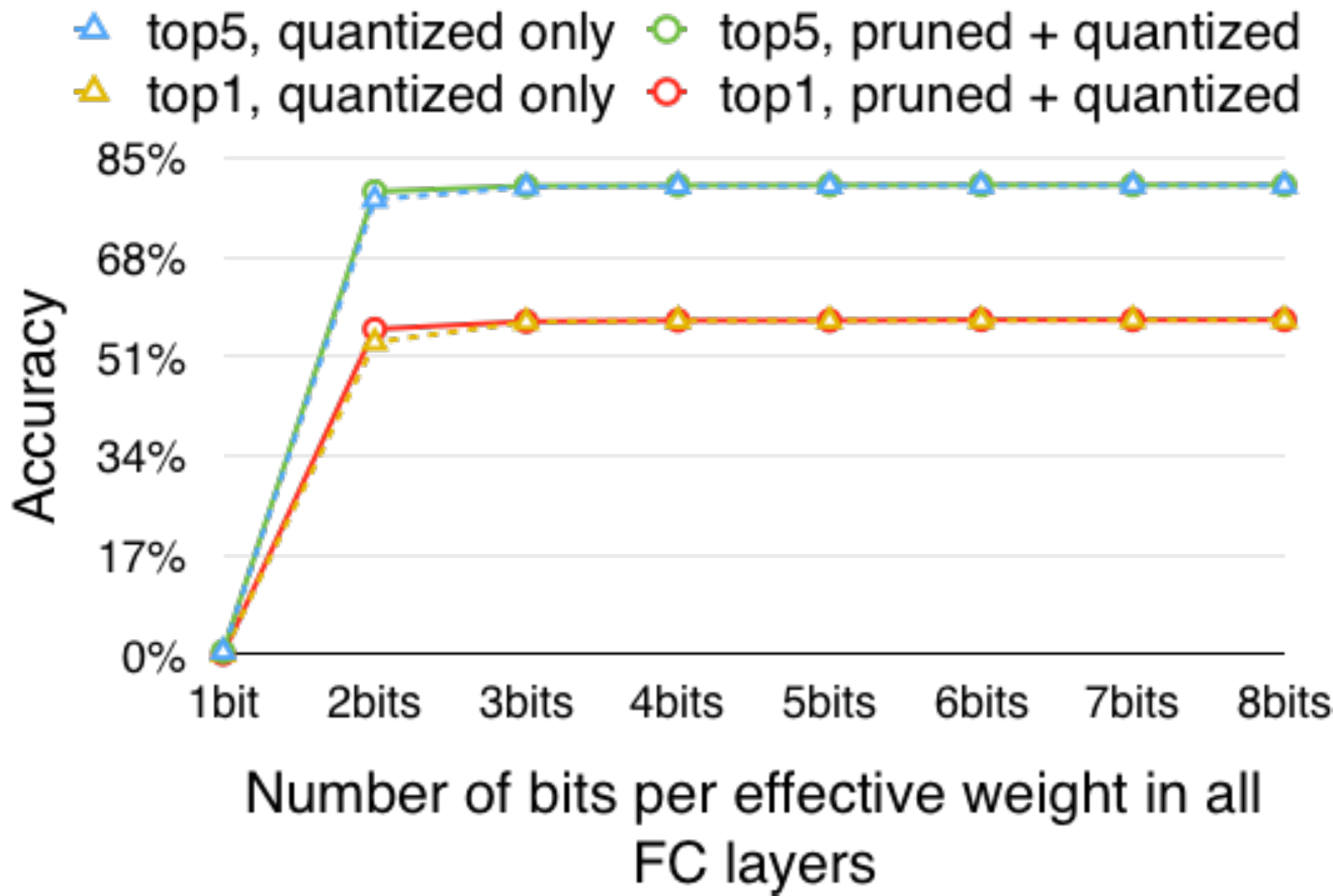}
% \hspace{-0.9em}
\includegraphics[width=0.329\textwidth]{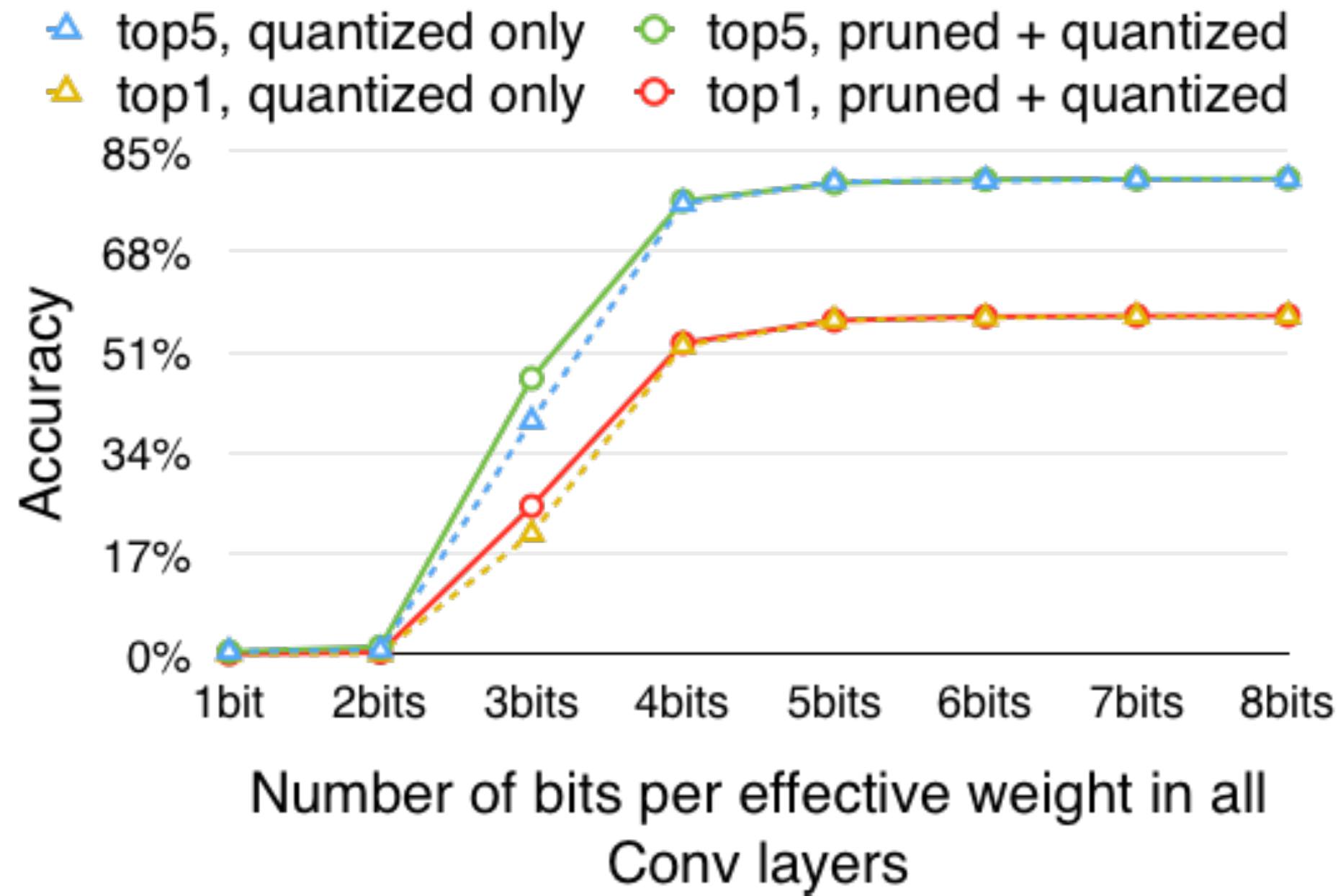}
% \hspace{-0.9em}
\includegraphics[width=0.329\textwidth]{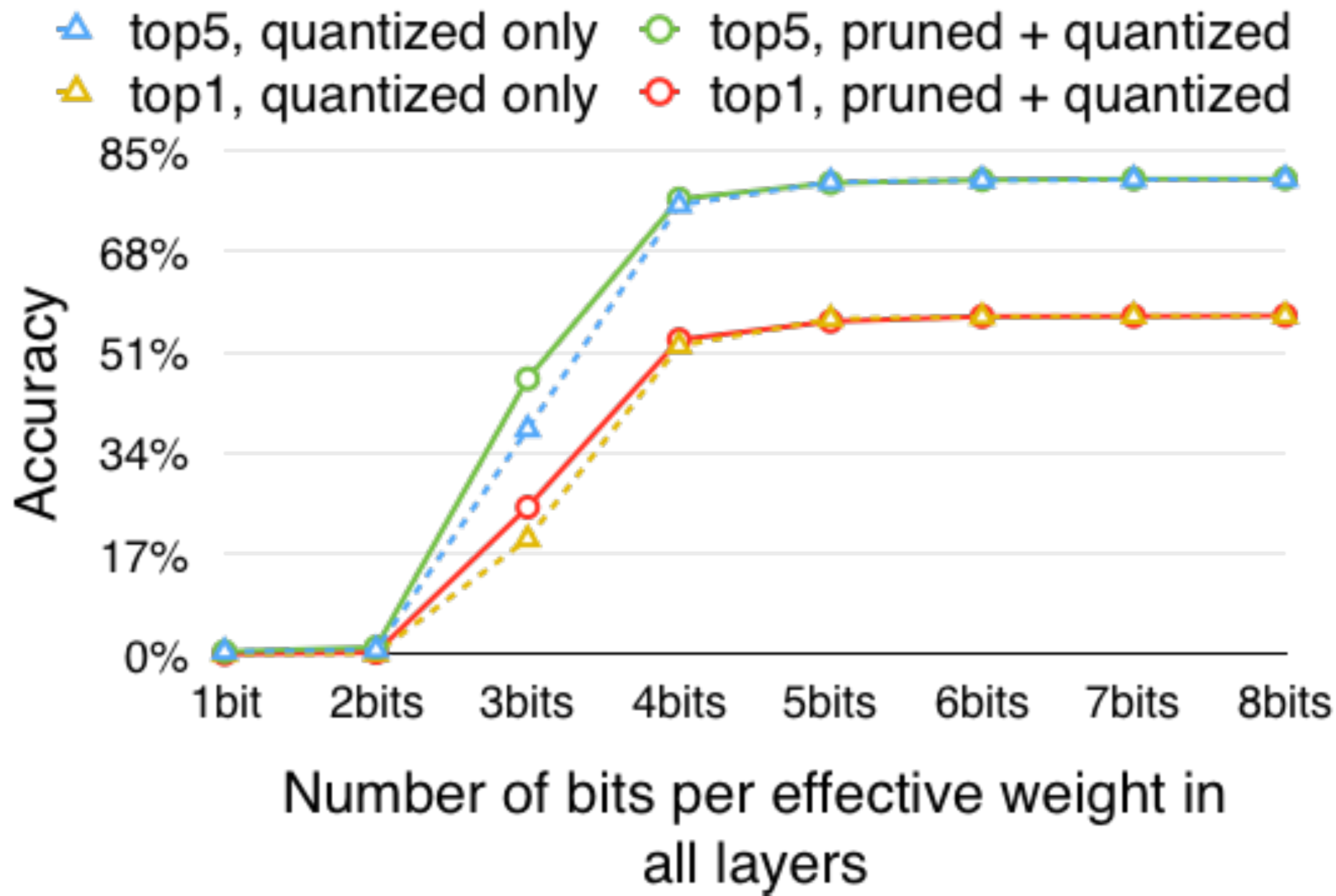}
\caption{Pruning doesn't hurt quantization. Dashed: quantization on  unpruned network. Solid: quantization on pruned network; Accuracy begins to drop at the same number of quantization bits whether or not the network has been pruned. Although pruning made the number of parameters less, quantization still works well, or even better(3 bits case on the left figure) as in the unpruned network.}
\vspace{-5pt}
\label{fig:prune_quantize}
\end{figure*}

\begin{figure*}[h!]
\centering
% \vspace{-30pt}
\includegraphics[width=0.49\textwidth]{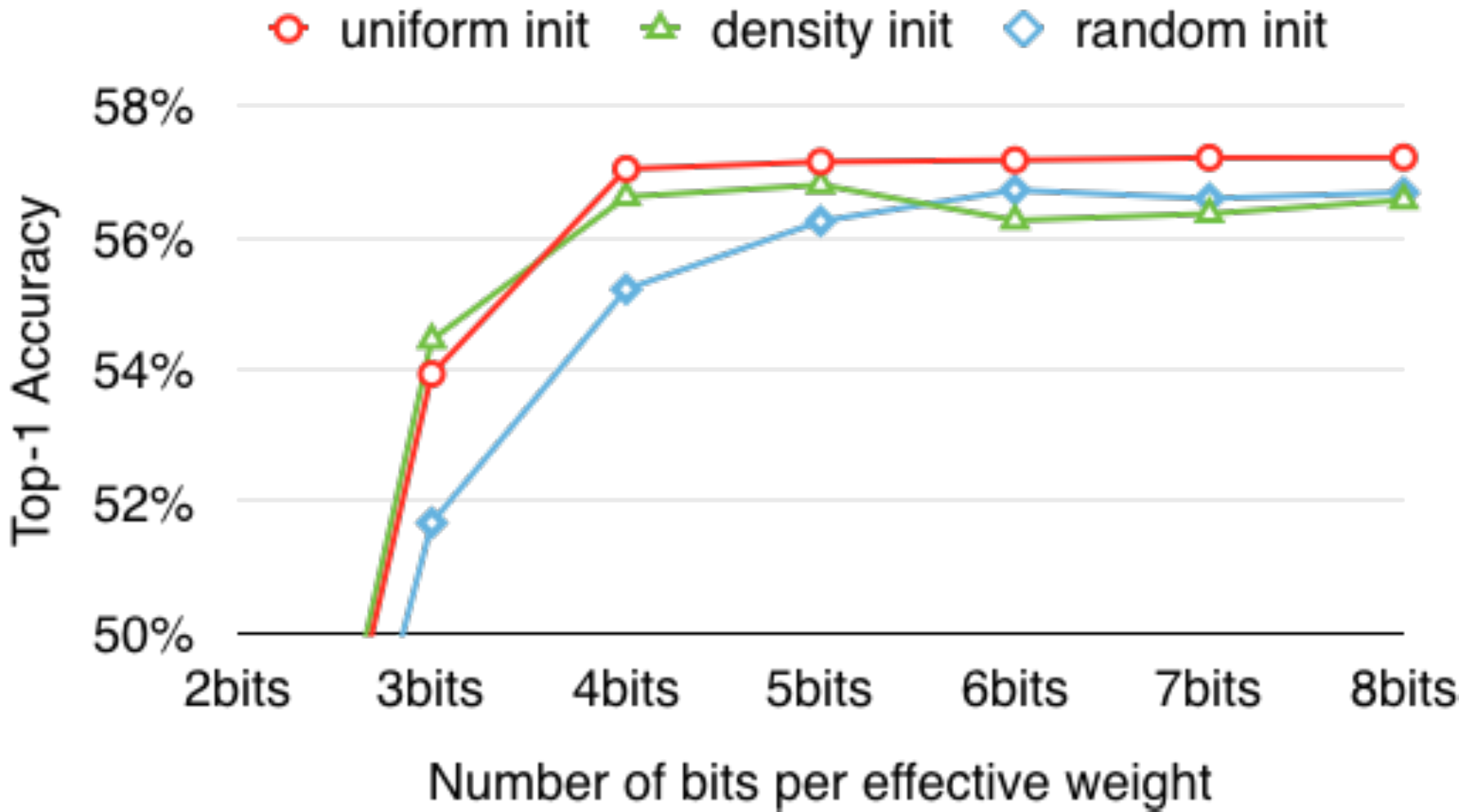}
% \hspace{1em}
\includegraphics[width=0.49\textwidth]{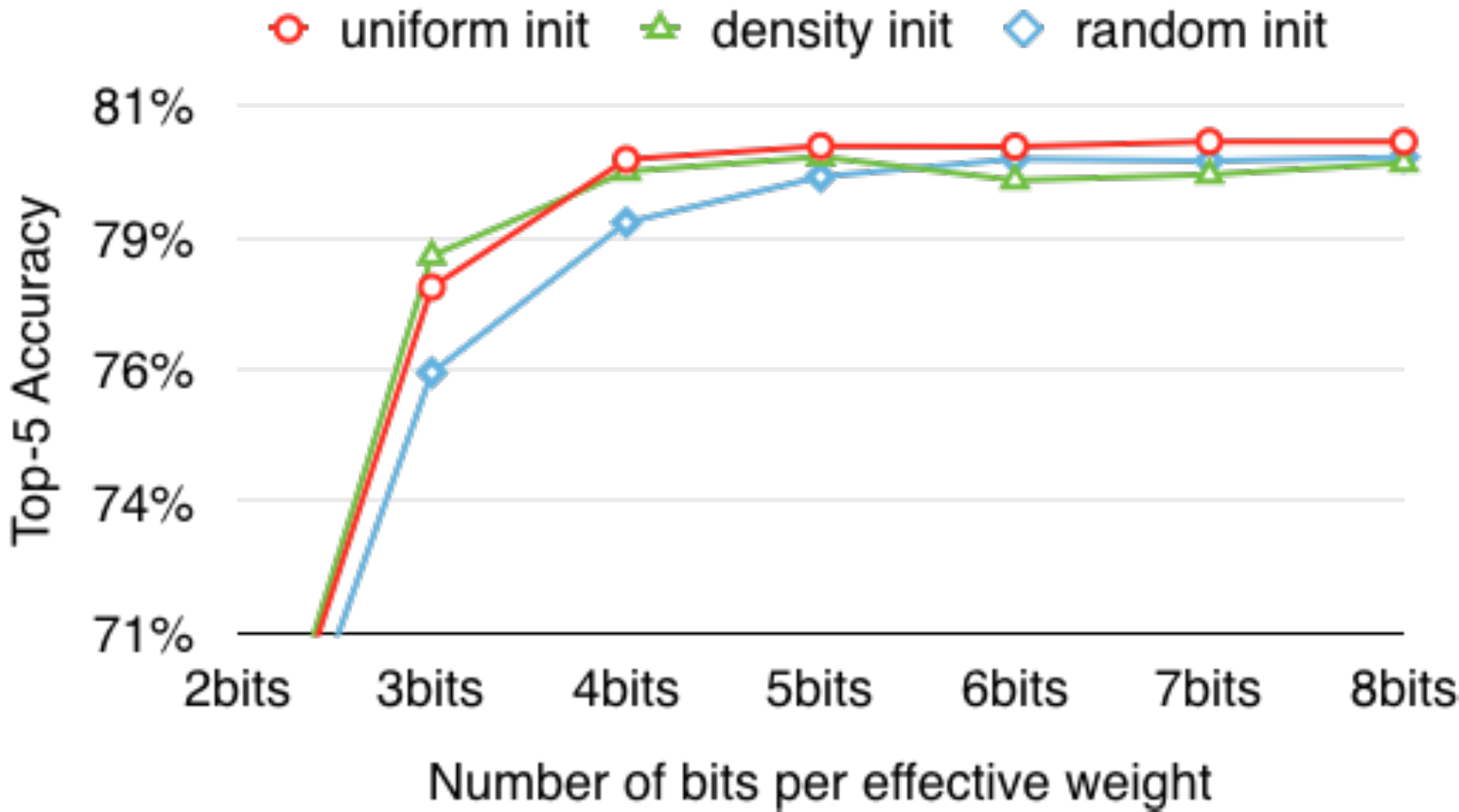}
\caption{Accuracy of different initialization methods. Left: top-1 accuracy. Right: top-5 accuracy. Linear initialization gives best result.}
\label{fig:init_acc}
\vspace{-10pt}
\end{figure*}

The three plots in Figure \ref{fig:prune_quantize} show how accuracy drops with fewer bits per connection for CONV layers (left), FC layers (middle) and all layers (right). Each plot reports both top-1 and top-5 accuracy. Dashed lines only applied quantization but without pruning; solid lines did both quantization and pruning. There is very little difference between the two. This shows that pruning works well with quantization.

Quantization works well on pruned network because unpruned AlexNet has 60 million  weights to quantize, while pruned AlexNet has only 6.7 million weights to quantize. Given the same amount of centroids, the latter has less error. 

The first two plots in Figure \ref{fig:prune_quantize} show that CONV layers require more bits of precision than FC layers. For CONV layers, accuracy drops significantly below 4 bits, while FC layer is more robust: not until 2 bits did the accuracy drop significantly.

%%%%%%%%%%%%%%%%%%%%%%  2. Centroid Initialization   %%%%%%%%%%%%%%%%%%%%%%

% \vspace{-3pt}
\subsection{Centroid Initialization} 
% \vspace{-3pt}

Figure \ref{fig:init_acc} compares the accuracy of the three different initialization methods with respect to top-1 accuracy (Left) and top-5 accuracy (Right). The network is quantized to  $2\sim 8$ bits as shown on x-axis.  Linear initialization outperforms the density initialization and random initialization in all cases except at 3 bits. 

The initial centroids of linear initialization spread equally across the x-axis, from the min value to the max value. That helps to maintain the large weights as the large weights play a more important role than smaller ones, which is also shown in network pruning \cite{song_pruning}. Neither random nor density-based initialization retains large centroids. With these initialization methods, large weights are clustered to the small centroids because there are few large weights. In contrast, linear initialization allows large weights a better chance to form a large centroid.

%%%%%%%%%%%%%%%%%%%%%%  Speedup  %%%%%%%%%%%%%%%%%%%%%%
\subsection{Speedup and Energy Efficiency}
\begin{figure}[t]
\centering
\vspace{-30pt}
\scalebox{1}[1]{\includegraphics[width=0.95\textwidth]{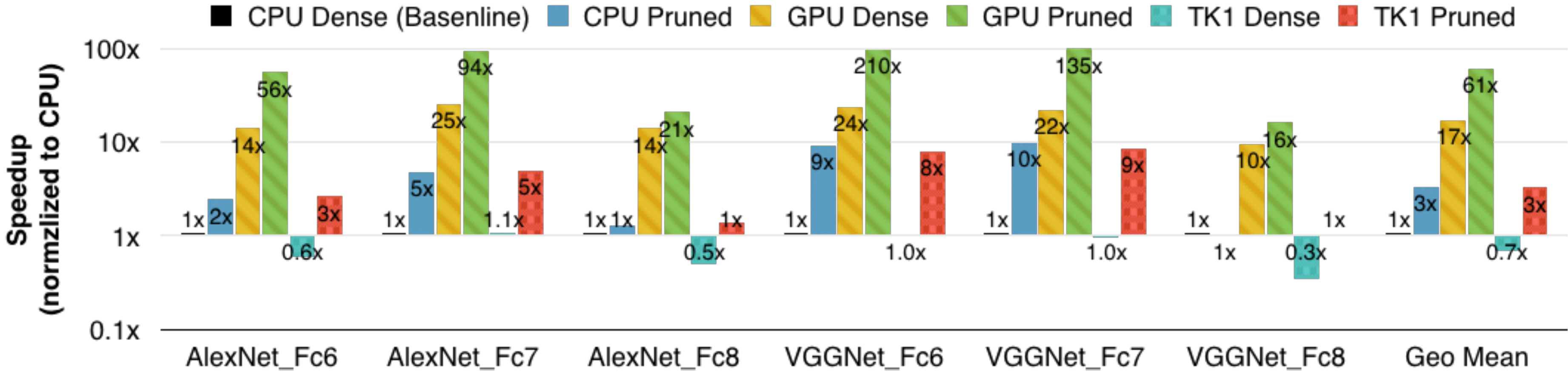}}
\caption{Compared with the original network, pruned network layer achieved $3\times$ speedup on CPU, $3.5\times$ on GPU and $4.2\times$ on mobile GPU on average. Batch size = 1 targeting real time processing. Performance number normalized to CPU.}
\label{fig:speedup}
\end{figure}

\begin{figure}[t]
\centering
\scalebox{1}[1]{\includegraphics[width=0.95\textwidth]{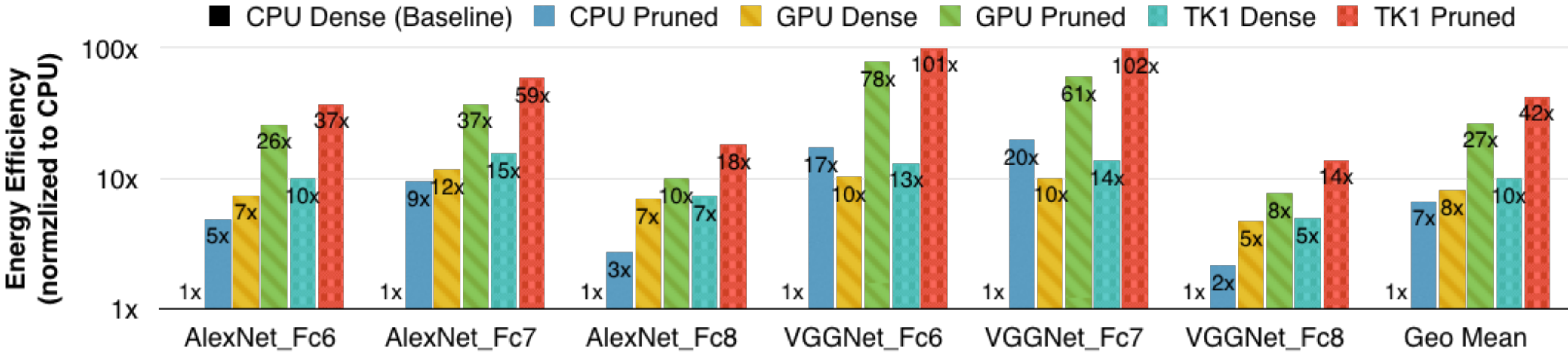}}
\caption{Compared with the original network, pruned network layer takes $7\times$ less energy on CPU, $3.3\times$ less on GPU and $4.2\times$ less on mobile GPU on average. Batch size = 1 targeting real time processing. Energy number normalized to CPU.}
\label{fig:efficiency}
\end{figure}

Deep Compression is targeting extremely latency-focused applications running on mobile, which requires real-time inference, such as pedestrian detection on an embedded processor inside an autonomous vehicle. Waiting for a batch to assemble significantly adds latency. So when bench-marking the performance and energy efficiency, we consider the case when batch size = 1. The cases of batching are given in Appendix~A.

Fully connected layer dominates the model size (more than $90\%$) and got compressed the most by Deep Compression~($96\%$ weights pruned in VGG-16). In state-of-the-art object detection algorithms such as fast R-CNN~\citep{girshick2015fast},    
upto $38\%$ computation time is consumed on FC layers on uncompressed model. So it's interesting to benchmark on FC layers, to see the effect of Deep Compression on performance and energy. Thus we setup our benchmark on FC6, FC7, FC8 layers of AlexNet and VGG-16. In the non-batched case, the activation matrix is a vector with just one column, so the computation boils down to dense / sparse matrix-vector multiplication for original / pruned model, respectively. Since current BLAS library on CPU and GPU doesn't support indirect look-up and relative indexing, we didn't benchmark the quantized model.

We compare three different off-the-shelf hardware: the NVIDIA GeForce GTX Titan X and the Intel Core i7 5930K as desktop processors (same package as NVIDIA Digits Dev Box) and NVIDIA Tegra K1 as mobile processor.  To run the benchmark on GPU, we used cuBLAS GEMV for the original dense layer. For the pruned sparse layer, we stored the sparse matrix in in CSR format, and used cuSPARSE CSRMV kernel, which is optimized for sparse matrix-vector multiplication on GPU. To run the benchmark on CPU, we used MKL CBLAS GEMV for the original dense model and MKL SPBLAS CSRMV for the pruned sparse model. 

To compare power consumption between different systems, it is important to measure power at a consistent manner~\citep{whitepaper}. For our analysis, we are comparing pre-regulation power of the entire application processor (AP) / SOC and DRAM combined. On CPU, the benchmark is running on single socket with a single Haswell-E class Core i7-5930K processor. CPU socket and DRAM power are as reported by the \texttt{pcm-power} utility provided by Intel. For GPU,  we used \texttt{nvidia-smi} utility to report the power of Titan X. For mobile GPU, we use a Jetson TK1 development board and measured the total power consumption with a power-meter. We assume $15\%$ AC to DC conversion loss, $85\%$ regulator efficiency and $15\%$ power consumed by peripheral components~\citep{TK1} to report the AP+DRAM power for Tegra K1.

\begin{table}[]
\centering
\vspace{-10pt}
\caption{Accuracy of AlexNet with different aggressiveness of weight sharing and quantization. 8/5 bit quantization has no loss of accuracy; 8/4 bit quantization, which is more hardware friendly, has negligible loss of accuracy of 0.01\%; To be really aggressive, 4/2 bit quantization resulted in 1.99\% and 2.60\% loss of accuracy.}
\label{my-label}
\begin{tabular}{c|cc|cc}
\#CONV bits / \#FC bits & Top-1 Error & Top-5 Error & \begin{tabular}[c]{@{}c@{}}Top-1 Error \\ Increase\end{tabular} & \begin{tabular}[c]{@{}c@{}}Top-5 Error \\ Increase\end{tabular} \\ \hline
 32bits /  32bits         & 42.78\%     & 19.73\%     & -                    & -                    \\
~8~bits / ~5~bits           & 42.78\%     & 19.70\%     & 0.00\%               & -0.03\%              \\
~8~bits / ~4~bits           & 42.79\%     & 19.73\%     & 0.01\%               & 0.00\%               \\
~4~bits / ~2~bits           & 44.77\%     & 22.33\%     & 1.99\%               & 2.60\%              
\end{tabular}
\vspace{-10pt}
\end{table}

The ratio of memory access over computation characteristic with and without batching is different. When the input activations are batched to a matrix the computation becomes matrix-matrix multiplication, where locality can be improved by blocking. Matrix could be blocked to fit in caches and reused efficiently. In this case, the amount of memory access is $O(n^2)$, and that of computation is $O(n^3)$, the ratio between memory access and computation is in the order of $1/n$. 

In real time processing when batching is not allowed, the input activation is a single vector and the computation is matrix-vector multiplication. In this case, the amount of memory access is $O(n^2)$, and the computation is $O(n^2)$, memory access and computation are of the same magnitude (as opposed to $1/n$). That indicates MV is more memory-bounded than MM. So reducing the memory footprint is critical for the non-batching case.

Figure~\ref{fig:speedup} illustrates the speedup of pruning on different hardware. There are 6 columns for each benchmark, showing the computation time of CPU~/~GPU~/~TK1 on dense~/~pruned network. Time is normalized to CPU.
When batch size = 1, pruned network layer obtained $3\times$ to $4\times$ speedup over the dense network on average because it has smaller memory footprint and alleviates the data transferring overhead, especially for large matrices that are unable to fit into the caches. For example VGG16's FC6 layer, the largest layer in our experiment, contains $25088 \times 4096  \times 4~Bytes \approx 400MB$ data, which is far from the capacity of L3 cache.

In those latency-tolerating applications , batching improves memory locality, where weights could be blocked and reused in matrix-matrix multiplication. In this scenario, pruned network no longer shows its advantage. We give detailed timing results in Appendix~A.

Figure~\ref{fig:efficiency} illustrates the energy efficiency of pruning on different hardware. We multiply power consumption with computation time to get energy consumption, then normalized to CPU to get energy efficiency. 
When batch size = 1, pruned network layer consumes $3\times$ to $7\times$ less energy over the dense network on average. Reported by \texttt{nvidia-smi}, GPU utilization is $99\%$ for both dense and sparse cases.

%%%%%%%%%%%%%%%%%%%%%%  3. Ratio   %%%%%%%%%%%%%%%%%%%%%%

\subsection{Ratio of Weights, Index and Codebook}
\begin{figure}[h!]
\centering
\scalebox{1}[1]{\includegraphics[width=0.8\textwidth]{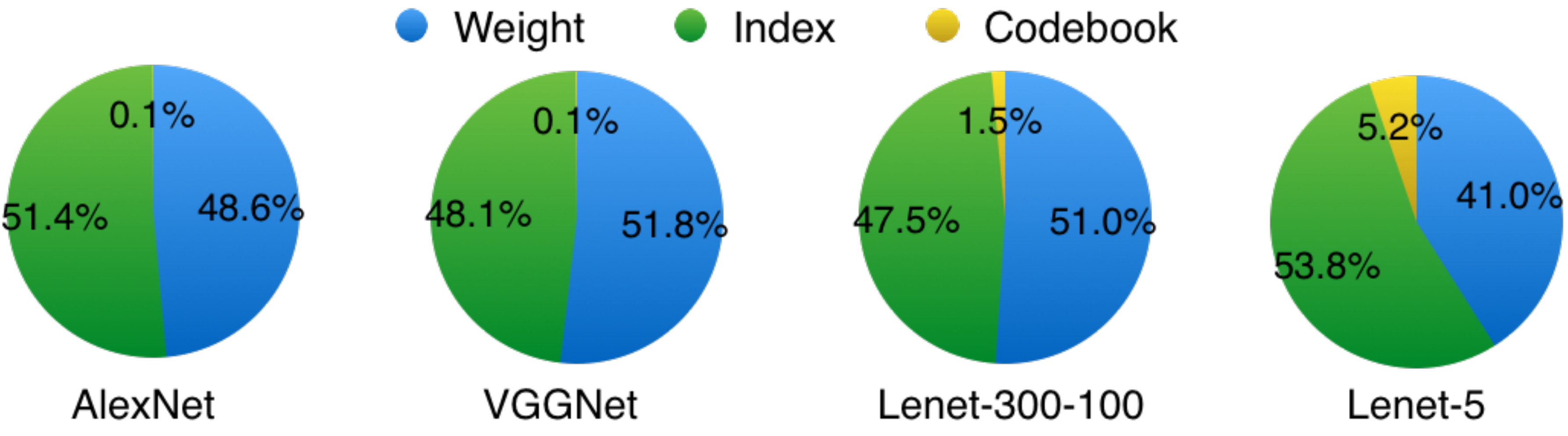}}
\caption{Storage ratio of weight, index and codebook.}
\label{fig:ratio}
\end{figure}

Pruning makes the weight matrix sparse, so extra space is needed to store the indexes of non-zero elements. Quantization adds storage for a codebook. The experiment section has already included these two factors. Figure \ref{fig:ratio} shows the breakdown of three different components when quantizing four networks. Since on average both the weights and the sparse indexes are encoded with 5 bits, their storage is roughly half and half. The overhead of codebook is very small and often negligible.

%%%%%%%%%%%%%%%%%%%%%%%%%%%%%%%%%%%%  Related Work  %%%%%%%%%%%%%%%%%%%%%%
\section{Related Work}
Neural networks are typically over-parametrized, and there is significant redundancy for deep learning models\citep{denil2013predicting}. This results in a waste of both computation and memory usage. There have been various proposals to remove the redundancy: \citet{vincentGoogle2011improving} explored a fixed-point implementation with 8-bit integer (vs 32-bit floating point) activations. \citet{hwang2014fixed} proposed an optimization method for the fixed-point network with ternary weights and 3-bit activations. \citet{anwar2015fixed} quantized the neural network using L2 error minimization and achieved better accuracy on MNIST and CIFAR-10 datasets.\citet{denton2014LinearStructure} exploited the linear structure of the neural network by finding an appropriate low-rank approximation of the parameters and  keeping the accuracy within 1\% of the original model. 

The empirical success in this paper is consistent with the theoretical study of random-like sparse networks with +1/0/-1 weights \citep{Arora14}, which have been proved to enjoy nice properties (e.g. reversibility), and to allow a provably polynomial time algorithm for training. 

Much work has been focused on binning the network parameters into buckets, and only the values in the buckets need to be stored. HashedNets\citep{chen2015compressing} reduce model sizes by using a hash function to randomly group connection weights, so that all connections within the same hash bucket share a single parameter value. In their method, the weight binning is pre-determined by the hash function, instead of being learned through training, which doesn't capture the nature of images. \citet{gong2014compressing} compressed deep convnets using vector quantization, which resulted in 1\% accuracy loss. Both methods studied only the fully connected layer, ignoring the convolutional layers.

\begin{table}[t]
\centering
\vspace{-30pt}
\caption{Comparison with other compression methods on AlexNet. \citep{collins2014memory} reduced the parameters by $4\times$ and with inferior accuracy. Deep Fried Convnets\citep {yang2014deep} worked on fully connected layers and reduced the parameters by less than $4\times$. SVD save parameters but suffers from large accuracy loss as much as 2\%. Network pruning \citep{song_pruning} reduced the parameters by $9\times$, not including index overhead. On other networks similar to AlexNet, \citep{denton2014LinearStructure} exploited linear structure of convnets and compressed the network by $2.4\times$ to $13.4\times$ layer wise, with 0.9\% accuracy loss on compressing a single layer. \citep{gong2014compressing} experimented with vector quantization and compressed the network by $16\times$ to $24\times$, incurring 1\% accuracy loss. }
\begin{tabular}{lllll}
\hline
\multicolumn{1}{l|}{Network}              & Top-1 Error& \multicolumn{1}{l|}{Top-5 Error} & \multicolumn{1}{l|}{Parameters}  & \begin{tabular}[c]{@{}l@{}}Compress\\ Rate\end{tabular} \\ \hline
\multicolumn{1}{l|}{Baseline Caffemodel \citep{caffemodel}}        & 42.78\%     & \multicolumn{1}{l|}{19.73\%}         & \multicolumn{1}{l|}{240MB}      & $1\times$ \\
\multicolumn{1}{l|}{Fastfood-32-AD \citep{yang2014deep}}       & 41.93\%     & \multicolumn{1}{l|}{-}         & \multicolumn{1}{l|}{131MB}      & $2\times$ \\
\multicolumn{1}{l|}{Fastfood-16-AD \citep{yang2014deep}}       & 42.90\%     & \multicolumn{1}{l|}{-}         & \multicolumn{1}{l|}{64MB}      & $3.7\times$ \\
\multicolumn{1}{l|}{Collins \& Kohli \citep{collins2014memory}}& 44.40\%     & \multicolumn{1}{l|}{-}         & \multicolumn{1}{l|}{61MB}      & $4\times$ \\
\multicolumn{1}{l|}{SVD \citep{denton2014LinearStructure}}                           & 44.02\%       & \multicolumn{1}{l|}{20.56\%}   & \multicolumn{1}{l|}{47.6MB}      & $5\times$ \\
\multicolumn{1}{l|}{Pruning \citep{song_pruning}}       & 42.77\%   & \multicolumn{1}{l|}{19.67\%}     & \multicolumn{1}{l|}{27MB}         & $9\times$     \\
\multicolumn{1}{l|}{Pruning+Quantization}     & 42.78\%   & \multicolumn{1}{l|}{19.70\%}     & \multicolumn{1}{l|}{8.9MB}         & $27\times$     \\
\multicolumn{1}{l|}{\bf{Pruning+Quantization+Huffman}}  & \bf{42.78}\%   & \multicolumn{1}{l|}{\bf{19.70\%}}     & \multicolumn{1}{l|}{\bf{6.9MB}}         & $\bf{35\times}$     \\ \hline
\end{tabular}

\label{table:compare}
\end{table}

There have been other attempts to reduce the number of parameters of neural networks by replacing the fully connected layer with global average pooling. The Network in Network architecture\citep{NiN} and GoogLenet\citep{szegedy2014GoogLenet} achieves state-of-the-art results on several benchmarks by adopting this idea. However, transfer learning, i.e. reusing features learned on the ImageNet dataset and applying them to new tasks by only fine-tuning the fully connected layers, is more difficult with this approach. This problem is noted by \citet{szegedy2014GoogLenet} and motivates them to add a linear layer on the top of their networks to enable transfer learning. 

Network pruning has been used both to reduce network complexity
and to reduce over-fitting.
An early approach to pruning was biased weight decay \citep{hanson1989comparing}.
Optimal Brain Damage \citep{lecun1989optimal} and Optimal Brain Surgeon \citep{hassibi1993second} prune networks to reduce the number of connections based on the Hessian of the loss function and suggest that such pruning is more accurate than magnitude-based pruning such as weight decay. A recent work \citep{song_pruning} successfully pruned several state of the art large scale networks and showed that the number of parameters could be reduce by an order of magnitude. There are also attempts to reduce the number of activations for both compression and acceleration \citet{van2015cross}.

\section{Future Work}
While the \emph{pruned} network has been benchmarked on various hardware, the \emph{quantized} network with weight sharing has not, because off-the-shelf cuSPARSE or MKL SPBLAS library does not support indirect matrix entry lookup, nor is the relative index in CSC or CSR format supported. So the full advantage of Deep Compression that fit the model in  cache is not fully unveiled. A software solution is to write customized GPU kernels that support this.  A hardware solution is to build custom ASIC architecture specialized to traverse the sparse and quantized network structure, which also supports customized quantization bit width. We expect this architecture to have energy dominated by on-chip SRAM access instead of off-chip DRAM access.

%%%%%%%%%%%%%%%%%%%%%%%%%%%%%%%%%%%%    Conclusion %%%%%%%%%%%%%%%%%%%%%%%%%%%%%%%%%%%%  
\section{Conclusion}

We have presented ``Deep Compression'' that compressed neural networks without affecting accuracy.
Our method operates by pruning the unimportant connections, quantizing the network using weight sharing, and then applying Huffman coding.
We highlight our experiments on AlexNet which reduced the weight storage by $35\times$ without loss of accuracy.  We show similar results for VGG-16 and LeNet networks compressed by $49\times$ and $39\times$ without loss of accuracy. This leads to smaller storage requirement of putting convnets into mobile app. After Deep Compression the size of these networks fit into on-chip SRAM cache (5pJ/access) rather than requiring off-chip DRAM memory (640pJ/access). This potentially makes deep neural networks more energy efficient to run on mobile. Our compression method also facilitates the use of complex neural networks in mobile applications where application size and download bandwidth are constrained.


\bibliography{iclr2016_conference}
\bibliographystyle{iclr2016_conference}

\newpage
\appendix 
\section{Appendix: detailed timing / power reports of dense \& sparse network layers}

\begin{table}[h!]
\centering
\caption{Average time on different layers. To avoid variance, we measured the time spent on each layer for 4096 input samples, and averaged the time regarding each input sample. For GPU, the time consumed by \texttt{cudaMalloc} and \texttt{cudaMemcpy} is not counted. For batch size = 1, \texttt{gemv} is used; For batch size = 64, \texttt{gemm} is used. For sparse case, \texttt{csrmv} and \texttt{csrmm} is used, respectively.}
\label{table:detailed_performance}
\begin{tabular}{p{2.4cm}l|llllll}
\hline
Time~(us) &                                                          & \begin{tabular}[c]{@{}l@{}}AlexNet\\ FC6\end{tabular} & \begin{tabular}[c]{@{}l@{}}AlexNet\\ FC7\end{tabular} & \begin{tabular}[c]{@{}l@{}}AlexNet\\ FC8\end{tabular} & \begin{tabular}[c]{@{}l@{}}VGG16\\ FC6\end{tabular} & \begin{tabular}[c]{@{}l@{}}VGG16\\ FC7\end{tabular} & \begin{tabular}[c]{@{}l@{}}VGG16\\ FC8\end{tabular} \\
\hline
\multirow{4}{*}{Titan X}                                                   & dense (batch=1)  & 541.5                                                 & 243.0                                                 & 80.5                                                  & 1467.8                                              & 243.0                                               & 80.5                                                \\
                                                                           & sparse (batch=1) & 134.8                                                 & 65.8                                                  & 54.6                                                  & 167.0                                               & 39.8                                                & 48.0                                                \\
                                                                           & dense (batch=64)  & 19.8                                                  & 8.9                                                   & 5.9                                                   & 53.6                                                & 8.9                                                 & 5.9                                                 \\
                                                                           & sparse (batch=64) & 94.6                                                  & 51.5                                                  & 23.2                                                  & 121.5                                               & 24.4                                                & 22.0                                                \\
\hline
\multirow{4}{*}{\begin{tabular}[c]{@{}l@{}}Core \\ i7-5930k\end{tabular}} & dense (batch=1)  & 7516.2                                                & 6187.1                                                & 1134.9                                                & 35022.8                                             & 5372.8                                              & 774.2                                               \\
                                                                           & sparse (batch=1) & 3066.5                                                & 1282.1                                                & 890.5                                                 & 3774.3                                              & 545.1                                               & 777.3                                               \\
                                                                           & dense (batch=64)  & 318.4                                                 & 188.9                                                 & 45.8                                                  & 1056.0                                              & 188.3                                               & 45.7                                                \\
                                                                           & sparse (batch=64) & 1417.6                                                & 682.1                                                 & 407.7                                                 & 1780.3                                              & 274.9                                               & 363.1                                               \\
\hline
\multirow{4}{*}{Tegra K1}                                                  & dense (batch=1)  & 12437.2                                               & 5765.0                                                & 2252.1                                                & 35427.0                                             & 5544.3                                              & 2243.1                                              \\
                                                                           & sparse (batch=1) & 2879.3                                                & 1256.5                                                & 837.0                                                 & 4377.2                                              & 626.3                                               & 745.1                                               \\
                                                                           & dense (batch=64)  & 1663.6                                                & 2056.8                                                & 298.0                                                 & 2001.4                                              & 2050.7                                              & 483.9                                               \\
                                                                           & sparse (batch=64) & 4003.9                                                & 1372.8                                                & 576.7                                                 & 8024.8                                              & 660.2                                               & 544.1   \\ 
                                            \hline
\end{tabular}
\end{table}

\vspace{10pt}
\begin{table}[h!]
\centering
\caption{Power consumption of different layers. We measured the Titan X GPU power with \texttt{nvidia-smi}, Core i7-5930k CPU power with \texttt{pcm-power} and Tegra K1 mobile GPU power with an external power meter (scaled to AP+DRAM, see paper discussion). During power measurement, we repeated each computation multiple times in order to get stable numbers.  On CPU, dense matrix multiplications consume $2x$ energy than sparse ones because it is accelerated with multi-threading. 
}
\label{tab:detailed_power}
\begin{tabular}{p{2.4cm}l|llllll}
\hline
Power~(Watts)                                                                   &           & \begin{tabular}[c]{@{}l@{}}AlexNet\\ FC6\end{tabular} & \begin{tabular}[c]{@{}l@{}}AlexNet\\ FC7\end{tabular} & \begin{tabular}[c]{@{}l@{}}AlexNet\\ FC8\end{tabular} & \begin{tabular}[c]{@{}l@{}}VGG16\\ FC6\end{tabular} & \begin{tabular}[c]{@{}l@{}}VGG16\\ FC7\end{tabular} & \begin{tabular}[c]{@{}l@{}}VGG16\\ FC8\end{tabular} \\
\hline
\multirow{4}{*}{TitanX}                                                    & dense (batch=1)  & 157                                                   & 159                                                   & 159                                                   & 166                                                 & 163                                                 & 159                                                 \\
                                                                           & sparse (batch=1) & 181                                                   & 183                                                   & 162                                                   & 189                                                 & 166                                                 & 162                                                 \\
                                                                           & dense (batch=64)  & 168                                                   & 173                                                   & 166                                                   & 173                                                 & 173                                                 & 167                                                 \\
                                                                           & sparse (batch=64) & 156                                                   & 158                                                   & 163                                                   & 160                                                 & 158                                                 & 161                                                 \\
\hline
\multirow{4}{*}{\begin{tabular}[c]{@{}l@{}}Core \\ i7-5930k\end{tabular}} & dense (batch=1)  & 83.5                                                  & 72.8                                                  & 77.6                                                  & 70.6                                                & 74.6                                                & 77.0                                                  \\
                                                                           & sparse (batch=1) & 42.3                                                  & 37.4                                                  & 36.5                                                  & 38.0                                                  & 37.4                                                & 36.0                                                  \\
                                                                           & dense (batch=64)  & 85.4                                                  & 84.7                                                  & 101.6                                                 & 83.1                                                & 97.1                                                & 87.5                                                \\
                                                                           & sparse (batch=64) & 37.2                                                  & 37.1                                                  & 38                                                    & 39.5                                                & 36.6                                                & 38.2                                                \\
\hline
\multirow{4}{*}{Tegra K1}                                                  & dense (batch=1)  & 5.1                                                   & 5.1                                                   & 5.4                                                   & 5.3                                                 & 5.3                                                 & 5.4                                                 \\
                                                                           & sparse (batch=1) & 5.9                                                   & 6.1                                                   & 5.8                                                   & 5.6                                                 & 6.3                                                 & 5.8                                                 \\
                                                                           & dense (batch=64)  & 5.6                                                   & 5.6                                                   & 6.3                                                   & 5.4                                                 & 5.6                                                 & 6.3                                                 \\
                                                                           & sparse (batch=64) & 5.0                                                     & 4.6                                                   & 5.1                                                   & 4.8                                                 & 4.7                                                 & 5.0    \\ 
                                                                           \hline
\end{tabular}
\end{table}

\end{document}